\documentclass[11pt]{article}

\usepackage{acl}

\usepackage{times}
\usepackage{latexsym}
\usepackage[T1]{fontenc}
\usepackage[utf8]{inputenc}
\usepackage{microtype}
\usepackage{inconsolata}
\usepackage{graphicx}

\usepackage{amsmath,amsfonts,bm}

\def\eqref#1{equation~\ref{#1}}

\def\1{\bm{1}}

\DeclareMathAlphabet{\mathsfit}{\encodingdefault}{\sfdefault}{m}{sl}
\SetMathAlphabet{\mathsfit}{bold}{\encodingdefault}{\sfdefault}{bx}{n}

\usepackage{amsthm}

\usepackage{hyperref}
\usepackage{url}
\usepackage{caption}
\usepackage{tabularx}
\usepackage[most]{tcolorbox}
\usepackage{xspace}
\usepackage{booktabs}   
\usepackage{multirow}  
\usepackage{makecell}   
\usepackage[ruled,vlined]{algorithm2e}
\SetKwInput{KwInput}{Input}
\SetKwInput{KwOutput}{Output}
\usepackage[table]{xcolor} 
\definecolor{grey}{rgb}{0.5,0.5,0.5}
\usepackage{wrapfig}   

\usepackage{graphicx}
\usepackage{float}
\usepackage{amsmath}
\usepackage{amssymb}
\usepackage{enumitem}
\usepackage{listings}
\usepackage{subcaption}

\lstdefinestyle{promptstyle}{
  basicstyle=\ttfamily\footnotesize,
  backgroundcolor=\color{gray!10},
  frame=single,
  breaklines=true,
  columns=fullflexible
}

\theoremstyle{definition}

\newcommand{\trqa}{\textsc{TSAQA}\xspace}

\title{\trqa: Time Series Analysis Question And Answering Benchmark}

\author{
\textbf{Baoyu Jing}$^{1*}$, 
\textbf{Sanhorn Chen}$^{1*}$, 
\textbf{Lecheng Zheng}$^2$, 
\textbf{Boyu Liu}$^1$, 
\textbf{Zihao Li}$^1$, 
\textbf{Jiaru Zou}$^1$, \\
\textbf{Tianxin Wei}$^1$,
\textbf{Zhining Liu}$^1$, 
\textbf{Zhichen Zeng}$^1$, 
\textbf{Ruizhong Qiu}$^1$, 
\textbf{Xiao Lin}$^1$, \\
\textbf{Yuchen Yan}$^3$, 
\textbf{Dongqi Fu}$^4$, 
\textbf{Jingchao Ni}$^5$, 
\textbf{Jingrui He}$^1$, 
\textbf{Hanghang Tong}$^1$ \\
$^1$University of Illinois at Urbana-Champaign, 
$^2$Virginia Polytechnic Institute and State University\\
$^3$Amazon, $^4$Meta AI, $^5$University of Houston \\
\{baoyuj2, sanhorn2, jingrui, htong\}@illinois.edu, 
lecheng@vt.edu,
jni7@uh.edu
}

\begin{document}
\maketitle

\begingroup
\renewcommand\thefootnote{*}
\footnotetext{Equal contribution.}
\endgroup

\begin{abstract}
Time series data are integral to applications across domains such as finance, healthcare, transportation, and environmental science.
While recent work has begun to explore time series question answering (QA), existing benchmarks still provide limited coverage of analytical capabilities under a standardized evaluation framework.
We introduce \trqa, a novel unified benchmark designed to broaden task coverage and evaluate diverse temporal analysis capabilities.
\trqa integrates 6 diverse tasks under a single framework ranging from \emph{conventional analysis}, including anomaly detection and classification, to \emph{advanced analysis}, such as characterization, comparison, data transformation, and temporal relationship analysis.
Spanning 210k samples across 13 domains, the dataset employs diverse formats, including \emph{true-or-false (TF)}, \emph{multiple-choice (MC)}, and a novel \emph{puzzling (PZ}), to comprehensively assess time series analysis.
Zero-shot evaluation shows that \trqa remains challenging for current Large Language Models (LLMs): best-performing commercial model, Gemini-2.5-Flash, achieves 65.08 average accuracy. Although instruction tuning improves open-source models' performance: the best-performing model, LLaMA-3.1-8B, shows significant room for improvement. We further evaluate language-capable time series foundation models (TSFMs), showing that \trqa extends beyond general-purpose LLMs. The data are available in \url{https://huggingface.co/datasets/TSAQA/TSAQA-Benchmark}.
\end{abstract}

\section{Introduction}
Effective analysis over temporal patterns of time series data is essential for real-world decision-making. Traditionally, research in time series has concentrated on a narrow set of tasks, notably forecasting, anomaly detection, imputation, and classification~\citep{torres2021deep, lim2021time, wen2022transformers}. 
While these problems have important applications, the scope of temporal analysis extends far beyond these settings.


Recent advances in Large Language Models (LLMs) have revolutionized natural language processing and multimodal learning~\citep{gpt4, llama3,gemma3,qwen3}.
This progress has inspired a growing interest in applying LLMs to time series analysis.
Early studies have explored leveraging LLMs for traditional time series tasks, such as forecasting and anomaly detection~\citep{zeng2023transformers,jin2023time,zhou2024can,zhang2024large}, leaving open to the question of whether LLMs can develop stronger temporal analysis abilities, such as understanding contextual information.

\begin{figure*}[t!]
    \centering
    \begin{subfigure}[t]{0.4\textwidth}
        \centering
        \includegraphics[width=\linewidth]{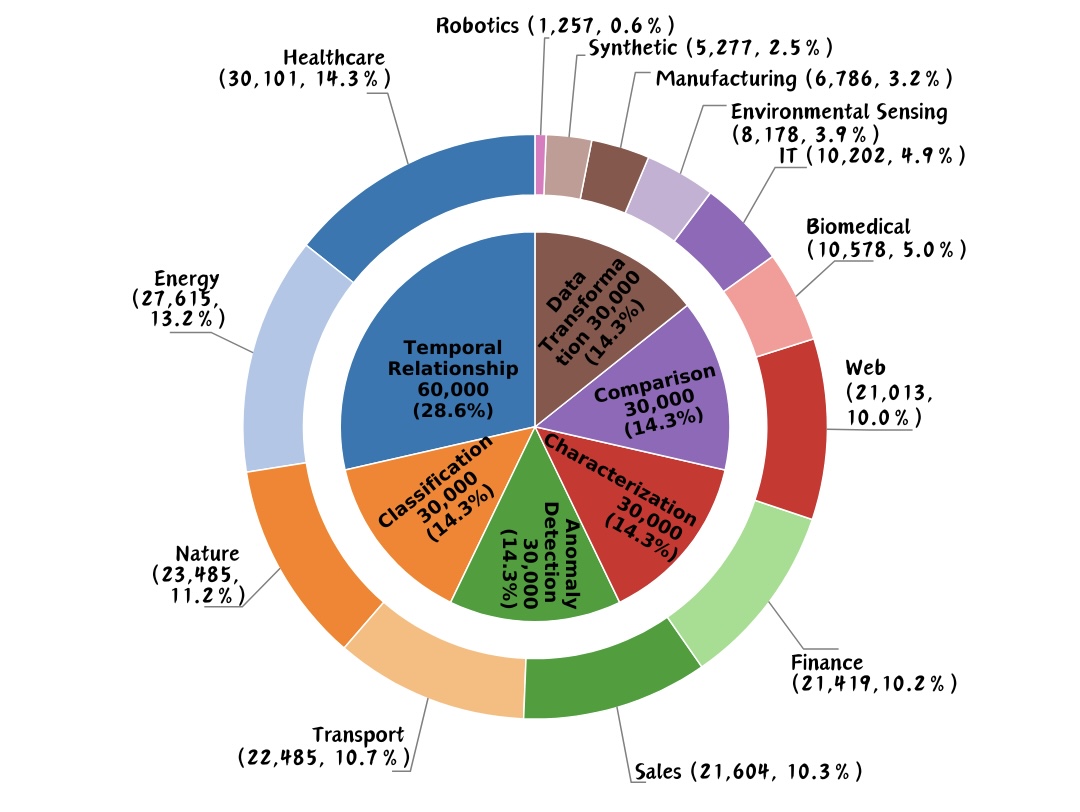}
        \caption{Domain and task distribution of \trqa.}
        \label{fig:domain_task_distribution}
    \end{subfigure}
    \begin{subfigure}[t]{0.58\textwidth}
        \centering
        \includegraphics[width=\linewidth]{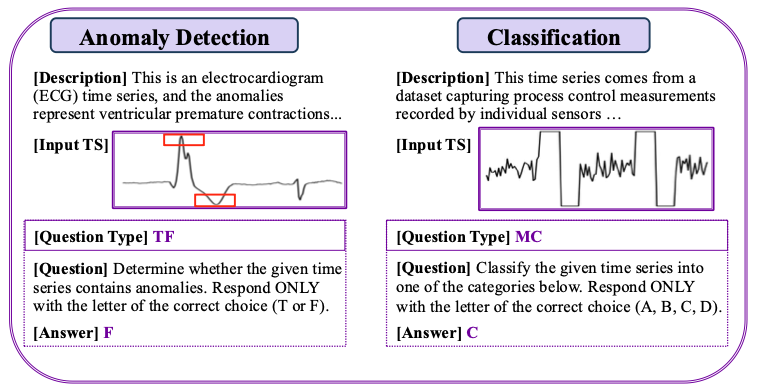}
        \caption{Illustration of conventional tasks.}
        \label{fig:conventional}
    \end{subfigure}
    \begin{subfigure}[t]{0.99\textwidth}
        \centering
    \includegraphics[width=\linewidth]{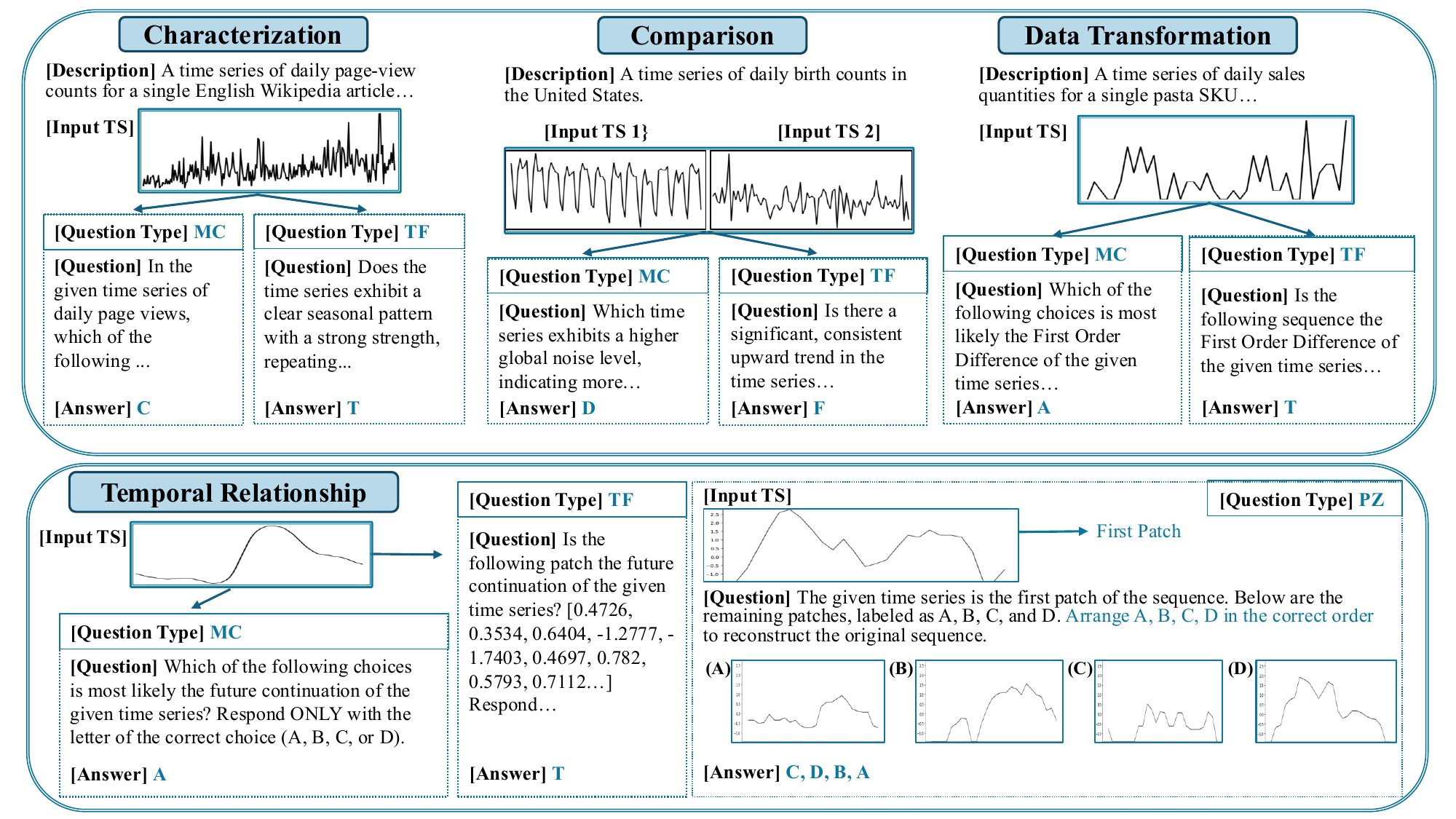}
        \caption{Illustration of advanced tasks.}
        \label{fig:advanced}
    \end{subfigure}
    \caption{Data distribution and tasks of \trqa.}
    \label{fig:figure1}
\end{figure*}

Time series question answering (QA) has recently emerged as a promising paradigm for pushing the boundaries of time series modeling beyond traditional tasks~\citep{DBLP:conf/emnlp/MerrillTGHA24, DBLP:conf/acl/UddinSHSS0CB25, DBLP:journals/corr/abs-2502-17812, DBLP:journals/corr/abs-2502-04395, DBLP:journals/corr/abs-2506-20093, DBLP:conf/acl/KongYHDZW0W25}.
By reformulating time series tasks through natural language queries, Time series QA enables models to tackle more complex questions about temporal patterns and dynamics, evaluating models' analytical capabilities.
For example, 
ITFormer~\citep{wang2025itformerbridgingtimeseries} introduces a \emph{domain-specific} EngineMT-QA dataset for aero engine time series. 
Time-MQA~\citep{DBLP:conf/acl/KongYHDZW0W25} constructs QA pairs that span both numeric tasks and open-ended QA tasks. While these efforts mark important progress, existing benchmarks remain fragmented in task coverage, modality, and evaluation design, limiting comprehensive assessment of temporal analytical capabilities. Moreover, questions requiring open-ended answers remain difficult to standardize objectively, complicating fair comparison across models.

In this paper, we introduce \trqa, Time Series Analysis Question and Answering Benchmark, a large-scale unified benchmark that addresses these limitations by covering diverse domains and tasks, while also providing standardized evaluation protocols.
Comparison between \trqa and existing datasets is provided in Table~\ref{tab:dataset_comparison}.
We curate and annotate 210k high-quality samples from 13 domains, as shown in Figure~\ref{fig:domain_task_distribution}.
\trqa integrates 6 distinct tasks grouped into two complementary categories:
(1) \emph{Conventional Analysis}: \emph{Anomaly detection} and \emph{Classification}.
(2) \emph{Advanced Analysis}: \emph{Characterization}, \emph{Comparison}, \emph{Data Transformation}, and \emph{Temporal Relationship}.
All tasks are cast into QA format with 3 question types: \emph{true-or-false (TF)}, \emph{multiple-choice (MC)}, and a novel \emph{puzzling (PZ)}. 

\trqa is designed to serve researchers studying temporal capabilities of language models, practitioners seeking references for time series QA, and the broader TSQA research community developing new models and methods. We carefully detail the \textit{data collection process}, \textit{benchmark construction}, \textit{dataset statistics}, and \textit{evaluation protocol}, ensuring transparency and reproducibility.
Our benchmark provides a standardized platform to evaluate various LLMs \citep{gpt4, claude, gemma3, llama3, qwen3} and TSFMs~\citep{wang2024chattimeunifiedmultimodaltime, xie2025chattsaligningtimeseries}. 

\begin{table*}[t!]
    \centering 
    \caption{Comparison of time series question answering datasets and benchmarks.}
    \label{tab:dataset_comparison} 
    \scriptsize
    \setlength{\tabcolsep}{3.5pt}
    {%
    \begin{tabular}{lcccrr}
        \toprule
        \textbf{Dataset} & \textbf{Tasks Scope} & \textbf{\# Analytical Tasks} & \textbf{\# Question Type} & \textbf{\# Domain} & \textbf{Size} \\
        \midrule
        TS-Insights \citep{zhang2023insight} & Captioning & 1 & 1 & 7 & 100k \\
        TSandLanguage \citep{DBLP:conf/emnlp/MerrillTGHA24} & Forecasting & 3 & 2 &10 & 230k \\
        CiK \citep{williams2025contextkeybenchmarkforecasting} &  Forecasting & 1 & 1 & 7 & 2.9k \\
        MTBench \citep{chen2025mtbenchmultimodaltimeseries} & Forecasting  & 4 & 3 & 2 & 42k \\
        \midrule
        TimeSeriesExam \citep{cai2024timeseriesexamtimeseriesunderstanding} & Various & 5 & 1 & 1 & 0.7k \\
        ChatTS \citep{xie2025chattsaligningtimeseries}  & Various    & 4 & 5 & 4 & 2.2k \\
        ITFormer \citep{wang2025itformerbridgingtimeseries} & Various    & 4 & 2 & 1 & 11k \\
        Time-MQA \citep{DBLP:conf/acl/KongYHDZW0W25}  & Various   &  5 & 4 & 12 & 200k\\
        SciTS \citep{wu2026scitsscientifictimeseries} & Various & 7 & 2 & 12 & 51k\\
        TSRBench \citep{yu2026tsrbenchcomprehensivemultitaskmultimodal} & Various & 15 & 1 & 14 & 4.1k\\
        MMTS-Bench \citep{yin2026mmtsbench} & Various & 4 & 2 & 5 & 2.4k\\
        TemporalBench \citep{weng2026temporalbenchbenchmarkevaluatingllmbased} & Various & 4 & 3 & 4 & 4.1k\\
        \hline
        \textbf{\trqa} (ours) & Various & \textbf{6} & \textbf{3} & \textbf{13} & \textbf{210k} \\
        \bottomrule
    \end{tabular}
    }
\end{table*}

In summary, our main contributions are threefold.
(1) We introduce \trqa, a novel large-scale benchmark comprising 210k samples across 13 domains, covering 6 tasks and 3 types of questions.
(2) We provide a detailed description of the benchmark’s construction along with comprehensive statistics.
(3) We conduct extensive evaluations of \trqa using a wide range of LLMs and TSFMs, accompanied by an in-depth analysis of their performance.

\section{Related Work}
\textbf{Time Series Analysis: From Numbers to Narratives.} 
Traditional research on time series has primarily focused on numerical sequences, enabling core tasks such as forecasting~\citep{torres2021deep}, imputation~\citep{wang2024deep}, and classification~\citep{mohammadi2024deep}, often treating them as isolated numeric signals~\citep{hamilton2020time}. In practice, however, time series are rarely independent of their surrounding context. They frequently interact with external information, such as textual reports or heterogeneous side signals, that shapes or enriches their dynamics~\citep{jiang2025multi, xu2025beyond, liu2025can, liu2024time, DBLP:journals/corr/abs-2502-08942}. Recognizing this, recent work has moved beyond purely numeric modeling to incorporate multimodal signals across domains including healthcare~\citep{johnson2016mimic, johnson2023mimic}, finance~\citep{li2024alphafin, dong2024fnspid}, retail~\citep{skenderi2024well}, and transportation~\citep{li2024urbangpt}. While much of this research leverages external modalities to improve numeric predictions on predefined tasks, a growing body of work instead positions \textit{natural language as a richer interface} for time series, using language as the medium for querying, reasoning, and interpreting temporal patterns~\citep{DBLP:conf/emnlp/MerrillTGHA24, williams2025contextkeybenchmarkforecasting, wang2025itformerbridgingtimeseries,  xie2025chattsaligningtimeseries, DBLP:conf/acl/KongYHDZW0W25}. Together, these efforts define the emerging direction of \textit{time series question answering}.

\noindent \textbf{Large Models on Time Series.} Advances in large language models (LLMs)~\citep{vaswani2017attention} have recently enabled general question answering over time series. A growing line of work integrates LLMs with time series for downstream tasks~\citep{chang2023llm4ts, alnegheimish2024large, yu2023temporal, jin2023time}, with extensions to multimodal language models as well~\citep{zhong2025time, DBLP:conf/emnlp/MerrillTGHA24, moon2022imu2clip}. Given their strong generalization ability, comprehensive evaluation is critical to ensure the transparency and reliability of LLMs in time series applications. 

Recent concurrent works~\citep{wu2026scitsscientifictimeseries, yin2026mmtsbench, yu2026tsrbenchcomprehensivemultitaskmultimodal, tsaia, weng2026temporalbenchbenchmarkevaluatingllmbased} have also expanded TSQA evaluation along complementary dimensions: SciTS~\citep{wu2026scitsscientifictimeseries} targets scientific domains with heterogeneous multivariate signals; TSRBench~\citep{yu2026tsrbenchcomprehensivemultitaskmultimodal} introduces multi-modal evaluation combining textual and visual chart inputs. Together, these works broaden TSQA along multiple axes, while \trqa emphasizes standardized large-scale analytical evaluation under a unified QA protocol.

\begin{table*}[t!]
\caption{Tasks of \trqa. TF, MC, and PZ denote true-or-false, multiple-choice, and puzzling.
}
\label{tab:task_summarization}
\scriptsize
\centering
\begin{tabular}{c|l|l|l}
\toprule
{\bf Group} & {\bf Task} & {\bf Description} & {\bf Question Type} \\
\midrule
\multirow{2}{*}{\centering\makecell[c]{Conventional\\ Analysis}}
& Anomaly Detection  & Determine whether the input contains anomalies. & TF \\
& Classification     & Classify the input time series.  & MC \\
\midrule
\multirow{4}{*}{\centering\makecell[c]{Advanced\\ Analysis}}
& Characterization       & Determine the characteristics of the time series. & TF \& MC \\
& Comparison             & Compare the characteristics of two time series.  & TF \& MC \\
& Data Transformation    & Identify the relationship between raw and transformed data. & TF \& MC \\
& Temporal Relationship  & Determine the temporal relationship of patches. & TF \& MC \& PZ\\
\bottomrule
\end{tabular}
\end{table*}

\section{\trqa Benchmark}
In this section, we introduce the proposed \trqa benchmark, which is designed to provide a benchmark for time series question answering. 
We begin by formulating the tasks and defining question types in Section~\ref{subsec:task_formulation}. 
Next, Section~\ref{subsec:data_collection} describes the data sources and preprocessing procedures. Section~\ref{subsec:benchmark_construction} then details the construction of the benchmark. Data statistics are discussed in Section~\ref{subsec:data_statistics}. Finally, Section~\ref{subsec:evaluation_protocol} outlines the evaluation protocols used to assess model performance.

\subsection{Task Formulation}
\label{subsec:task_formulation}

\textbf{Task Taxonomy.}
As shown in Table~\ref{tab:task_summarization} and Figure~\ref{fig:figure1}, the proposed \trqa benchmark encompasses two groups of tasks with 6 diverse tasks. We intentionally designed it to span a spectrum from fundamental analytical properties to more complex structural and relational reasoning, enabling evaluation across levels of temporal understanding.

The first group, 
\emph{Conventional Analysis}, includes fundamental tasks in time series analysis:
(1) \textit{Anomaly Detection} - identifies irregular or unexpected patterns in time series; 
(2) \textit{Classification} - recognizes the distinguishable semantic pattern of a time series that is pertinent to a class.
The second group, \emph{Advanced Analysis}, consists of novel analytical tasks about intrinsic properties of time series:
(3) \textit{Characterization} - infers fundamental properties such as trend, seasonality, and dispersion; 
(4) \textit{Comparison} - analyzes relative similarities and differences between two time series; 
(5) \textit{Data Transformation}- understands relationships between original and transformed time series, e.g., Fourier transform; 
and (6) \textit{Temporal Relationship} - captures the chronological dependencies among time series patches.
These advanced tasks push the boundaries of conventional time series modeling, fostering models to grasp cognitive concepts of time series and analyze over human questions.

To bring all tasks under a single umbrella, we formulate them in a unified QA format. 
Every instance is converted into a time series input $X$ paired with contextual information $C$ and a question $Q$, and the model is expected to provide the correct answer $A$, 
where $C$ and $Q$ are expressed by natural language.
Let $f$ denote the model, then the \trqa problem is formulated as:
\begin{equation}
    A = f(X, C, Q).
\end{equation}
\textbf{Question Types.}
\trqa benchmark encompasses a wide variety of question types, such as \emph{true-or-false (TF)}, \emph{multiple-choice (MC)}, and \emph{puzzling (PZ)} questions. 
A \textit{TF} question requires the model to determine whether a claim about the input time series is True (T) or False (F).
A \textit{MC} question requires the model to select the correct claim about the input.
In addition, we introduce a novel \emph{puzzling (PZ)} question, which is valuable because it represents realistic, human-like problem settings \cite{fissler2018jigsaw} and were shown to 
be effective in evaluating models’ general cognitive abilities, as demonstrated in computer vision \cite{noroozi2016unsupervised}.
In this question, models are given the first patch of a time series, along with the remaining shuffled patches, and is instructed to correct their chronological order. Together, the 3 closed-ended question types enable objective and reproducible evaluation across models.

\subsection{Data Collection}
\label{subsec:data_collection}
We collect and preprocess time series data from diverse public sources to ensure broad coverage and representativeness.
At its center are the core datasets, which serve as the foundation for most tasks. 
In addition, the benchmark integrates two specialized sources: classification datasets and anomaly detection datasets. 
This subsection describes these data sources and the selection criteria.
More details are presented in Appendix \ref{appendix:data_collection}.

\noindent \textbf{Core Datasets.}
We collect high-quality real-world time series data from a wide range of domains that are used by time series foundation model benchmarks, such as Lotsa~\citep{woo2024unified}, Time-300B~\citep{shi2024timemoe}, and UTSD~\citep{ma2024utsd}.
To ensure data quality, we retain only sequences with a minimum length of 1k. 
We further filter sequences with a missing rate greater than 1\% or an outlier rate (the proportion of points lying beyond three times the interquartile range (3×IQR)) exceeding 5\%.
More details are presented in Appendix~\ref{appendix:core_dataset}

\noindent \textbf{Anomaly Detection Datasets.}
We extract data from multiple time-series anomaly detection benchmarks~\citep{paparrizos2022tsb, DBLP:conf/kdd/SuZNLSP19}, including ECG~\citep{moody2001impact}, SMD~\citep{DBLP:conf/kdd/SuZNLSP19}, MGAB~\citep{thill10markusthill} Genesis~\citep{von2018anomaly}, GHL~\citep{DBLP:journals/corr/FilonovLV16}, Occupancy~\citep{candanedo2016accurate}.
These datasets span various domains.
More details are presented in Appendix~\ref{appendix:anomaly_detection_dataset}.

\noindent \textbf{Classification Datasets.}
Our classification data comes from the univariate UCR Archive~\citep{dau2019ucr}. We select datasets with at most four classes and sequence lengths under 400, and enrich them with textual descriptions from the official documentation. The resulting subset spans diverse domains. More details are presented in Appendix~\ref{appendix:classification_dataset}.

\subsection{Benchmark Construction}
\label{subsec:benchmark_construction}
In this subsection, we describe the construction of the benchmark for each task.
To maintain balance across tasks, we allocate an equal number of samples (30k) to each task, except for the temporal relationship task, which we allocate 60k samples since \textit{PZ} is very challenging. Apart from classification and anomaly detection, samples for all other tasks are drawn from the \textit{Core Datasets} (Section~\ref{subsec:data_collection}) using \textit{Hierarchical Random Sampling} (Algorithm~\ref{algo:hierarchical_random_sampling}) to ensure a balanced distribution across domains, datasets, and sequences.
Unless otherwise specified, 
each sample has a random length within $[32,256]$, fixed decimal levels, and is z-scored normalized before input to reduce data bias (Appendix~\ref{appendix:data-bias}).

Finally, each task’s samples are randomly partitioned into 70\% for training, 10\% for validation, and 20\% for testing.
Next, we describe the construction process for each task.
More details and examples can be found in Appendix~\ref{appendix:benchmark_construction}-\ref{appendix:examples}.

\noindent \textbf{Data Transformation.}
This task evaluates the model’s ability to infer the transformation relationship between the input time series and its transformed counterpart, which is generated from one of  Fourier transform, wavelet transform, or first-order differencing.
We then use predefined templates to formulate the task as either \textit{TF} or \textit{MC} questions.
In \textit{TF} questions, the model is asked to determine whether a given sequence is the correct transformation (e.g., the results of the Fourier transform) of the input time series $\mathbf{x}$.
In \textit{MC} questions, the model is required to select the correct transformed sequence given the input time series $\mathbf{x}$ and the specified transform operation (e.g., Fourier transform).
All transformations are computed using professional libraries~\citep{harris2020array,virtanen2020scipy}.
The correct transformation is computed directly from the input $\mathbf{x}$, whereas incorrect transformations are generated from other randomly sampled time series $\mathbf{x}'$.
For more details, please refer to Appendix~\ref{appendix:data_transformation}.

\noindent \textbf{Temporal Relationship.}
This task evaluates the model’s ability to infer the temporal structure among time series patches, testing 3 core capabilities: \textit{Structural Continuity}, \textit{Chronological Reasoning}, and \textit{Contextual Discrimination}.
This task is formulated as \textit{TF}, \textit{MC}, or \textit{PZ} questions.
Given the first chronological patch $\mathbf{x}$, a \textit{TF} question asks the model to determine whether a candidate patch $\mathbf{y}$ is the immediate successor of $\mathbf{x}$,
while an \textit{MC} question asks the model to choose the correct next patch from candidates $[\mathbf{y}_1,\mathbf{y}_2,\mathbf{y}_3,\mathbf{y}_4]$.
The false candidates are randomly sampled from the full dataset, but from sequences different from that of $\mathbf{x}$.
A \textit{PZ} question presents four shuffled successor patches of $\mathbf{x}$ and asks the model to arrange them in the correct chronological order.
All questions are generated using predefined templates.
See Appendix~\ref{appendix:temporal_relationship} for more details.

\noindent \textbf{Characterization.}
The characterization task assesses the model’s capability to analyze fundamental properties of time series, including trend, seasonality, and dispersion.
Questions are posed as \textit{TF} or \textit{MC}, and final answers are determined through multi-LLM consensus. Given a sample $\mathbf{x}$ and its meta data, we first instruct GPT-4o~\citep{hurst2024gpt} to generate question and answer pairs based on a randomly selected subset of one to three topics (from Table~\ref{table:topics}), and question type (\textit{TF} or \textit{MC}).

Briefly,  the process involves the following steps:
(1) We instruct GPT to generate captions for the input and randomly select a sub-topic for each topic (e.g., selecting the sub-topic “trend direction” under the topic “trend”);
(2) GPT is instructed to generate a QA pair based on the inputs, captions, sub-topics, and the specified question type;
(3) GPT performs a self-check of the generated QA pair and provides a confidence score~\citep{tian-etal-2023-just}, where only QA pair with a high confidence is retained;
(4) We further leverage other powerful LLMs, including GPT-4.1, Gemini-2.5-Flash, and Claude-3.5-Sonnet, along with the answer given by GPT-4o to produce a consensus answer and reduce model bias. More details in Appendix~\ref{appendix:characterization}.

\noindent \textbf{Comparison.}
The comparison task assesses the model’s ability to analyze the relative characteristics of the two time series, such as shape and correlation.
Similar to the characterization task, this task is also formulated as \textit{TF} or \textit{MC} questions.
We first obtain an anchor sample $\mathbf{x}$ from domain $M$, dataset $D$, and sequence $S$.
Given the anchor $\mathbf{x}$, we construct a set of 10 comparison samples $\{\mathbf{x}'_1\dots\mathbf{x}'_{10}\}$ with the same length as $\mathbf{x}$.
Among which, one is drawn from the sequence $S$, two from other sequences within dataset $D$, three from other datasets within domain $M$, and four from other domains.
We also use a process similar to the characterization task to generate QA pairs.
More details can be found in Appendix~\ref{appendix:comparison}.

\begin{table*}[t!]
    \centering
    \caption{Main results. A.D. denotes anomaly detection, CLS denotes classification. MC, TF, and PZ denote multiple-choice, true-or-false, and puzzling, respectively. SFT stands for supervised fine-tuning. The best and second-best results are highlighted in \textbf{bold} and \underline{underlined}, respectively.}
    \label{tab:main_results}
    \scriptsize
    \setlength{\tabcolsep}{3.2pt}
    \begin{tabular}{c|l |c| c| c c| c c| c c| c c c | c}
        \toprule
        \multirow{2}{*}{\textbf{Group}} &
        \textbf{Task} & 
        \multicolumn{1}{c|}{\textbf{A.D.}} & 
        \multicolumn{1}{c|}{\textbf{CLS}} & 
        \multicolumn{2}{c|}{\textbf{Characterization}} & 
        \multicolumn{2}{c|}{\textbf{Comparison}} & 
        \multicolumn{2}{c|}{\textbf{Data Transform}} & 
        \multicolumn{3}{c|}{\textbf{Temporal Relation}} & 
        \multirow{2}{*}{\textbf{Overall}}\\
        \cline{2-13}
        & {Question Type} & TF & MC & TF & MC & TF & MC & TF & MC & TF & MC & PZ \\
        \midrule

        \multirow{10}{*}{\makecell{Zero\\Shot}}
        & GPT-4.1       & 55.85 & 50.38 & \textbf{92.97} & \textbf{89.36} & 83.57 & 76.99 & 54.36 & 51.13 & 65.90 & 79.09 & 45.77 & 62.82\\
        & GPT-4o        & 54.32 & 47.20 & 88.15 & 84.15 & 78.61 & 69.07 & 60.66 & 53.24 & 62.25 & 75.58 & 45.61 & 60.73\\
        & Claude-3.5-Sonnet & 51.27 & 41.23 & 74.39 & 78.45 & 66.59 & 74.14 & 65.79 & 57.07 & 82.05 & 82.15 & 54.56 & 61.19\\
        & Gemini-2.5-Flash  & 52.08 & 49.07 & 85.48 & 81.08 & 77.79 & 72.21 & 63.62 & 60.17 & 75.05 & 84.49 & 60.84 & 65.08\\
        \cline{2-14}
        & Qwen3-8B          & 50.60 & 50.52 & 77.35 & 66.87 & 71.04 & 63.21 & 52.43 & 34.46 & 65.22 & 67.14 & 21.93 & 51.04\\
        & LLaMA3.1-8B       & 54.92 & 50.20 & 68.10 & 62.26 & 67.84 & 49.98 & 51.90 & 36.56 & 54.82 & 40.95 &  6.80 & 44.93\\
        & Ministral-8B      & 53.35 & 34.08 & 71.06 & 63.93 & 47.54 & 52.90 & 50.70 & 25.28 & 50.58 & 33.88 & 30.77 & 44.65\\
        \cline{2-14}
        & Qwen3-0.6B       & 50.40 & 35.83 & 62.00 & 48.78 & 58.03 & 37.51 & 49.03 & 23.62 & 51.99 & 37.33 & 13.38 & 39.06\\
        & LLaMA3.2-1B      & 49.47 & 39.48 & 63.74 & 52.55 & 61.02 & 36.82 & 48.87 &  4.20 & 48.97 &  5.44 &  6.76 & 35.70\\
        & Gemma3-1B        & 49.15 & 49.83 & 63.74 & 47.71 & 61.19 & 43.37 & 49.37 & 24.88 & 49.42 & 25.84 & 23.97 & 43.03\\
        \midrule
        \multirow{8}{*}{\makecell{Instruction\\Tuning}}
        & Qwen3-8B         & \underline{87.70} & \underline{90.05} & 92.37 & \underline{85.42} & \underline{86.55} & \underline{79.08} & \underline{89.84} & \underline{84.99} & \underline{96.84} & \textbf{97.56} & \underline{66.21} & \underline{84.29}\\
        & LLaMA3.1-8B      & \textbf{91.02} & \textbf{91.27} & \underline{92.44} & 83.68 & \textbf{86.72} & \textbf{79.31} & \textbf{90.17} & \textbf{86.62} & \textbf{96.94} & \underline{97.41} & \textbf{67.68} & \textbf{85.26}\\
        & Ministral-8B     & 71.56 & 74.28 & 91.31 & 80.78 & 84.14 & 74.63 & 75.15 & 71.61 & 94.07 & 94.15 & 56.82 & 74.74\\
        \cline{2-14}
        & Qwen3-0.6B       & 83.68 & 85.78 & 89.38 & 74.87 & 80.65 & 64.84 & 80.51 & 73.28 & 93.92 & 93.79 & 63.34 & 78.32\\
        & LLaMA3.2-1B      & 83.08 & 83.83 & 87.71 & 74.37 & 78.61 & 60.88 & 68.09 & 51.67 & 91.39 & 88.81 & 57.53 & 73.48\\
        & Gemma3-1B        & 83.10 & 84.05 & 87.88 & 72.54 & 78.61 & 59.31 & 64.06 & 45.23 & 91.00 & 88.05 & 42.92 & 69.70\\
        \cline{2-14}
        & ChatTime & 70.43 & 68.98 & 85.15 & 60.33 & 72.87 & 52.64 & 69.69 & 51.27 & 72.18 & 61.90 & 4.58 & 55.19 \\
        & ChatTS & 12.17 & 33.07 & 58.08 & 43.58 & 20.24 & 45.47 & 13.52 & 27.62 & 38.33 & 72.38 & 18.17 & 30.10 \\
        \bottomrule
    \end{tabular}
\end{table*}

\noindent \textbf{Anomaly Detection.}
The anomaly detection task evaluates the model’s ability to recognize anomalous patterns in the input time series, which is formulated as a \textit{TF} question. Each sample $\mathbf{x}$ is randomly cropped from a full sequence of the anomaly detection dataset. Since anomalous samples are much fewer than normal ones, we downsample the normal samples to balance the classes at a 1:1 ratio. The questions are composed using a predefined template. See Appendix~\ref{appendix:anomaly_detection} for more details.

\noindent \textbf{Classification.}
The classification task evaluates the model’s ability to categorize input time series based on their patterns and characteristics. We reformulate the classification task into the \textit{MC} question format, where the original numeric class, labels, e.g., 0 or 1, are converted into informative textual choices, e.g., ``Cabernet Sauvignon'' or ``Shiraz''.
See Appendix \ref{appendix:classification} for more details.

\subsection{Data Statistics}
\label{subsec:data_statistics}

\begin{figure*}[t!]
    \centering
    \begin{subfigure}[t]{0.25\textwidth}
        \centering
        \includegraphics[width=\linewidth]{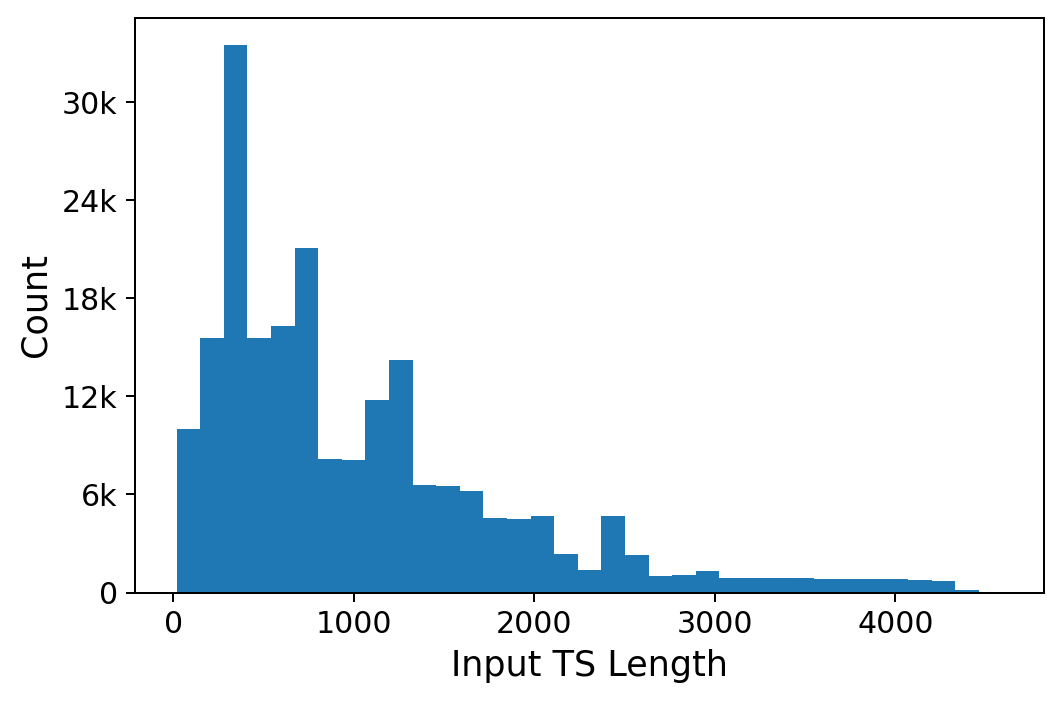}
        \caption{Time series length.}
        \label{fig:hist_ts_len}
    \end{subfigure}
    \hfill
    \begin{subfigure}[t]{0.25\textwidth}
        \centering
        \includegraphics[width=\linewidth]{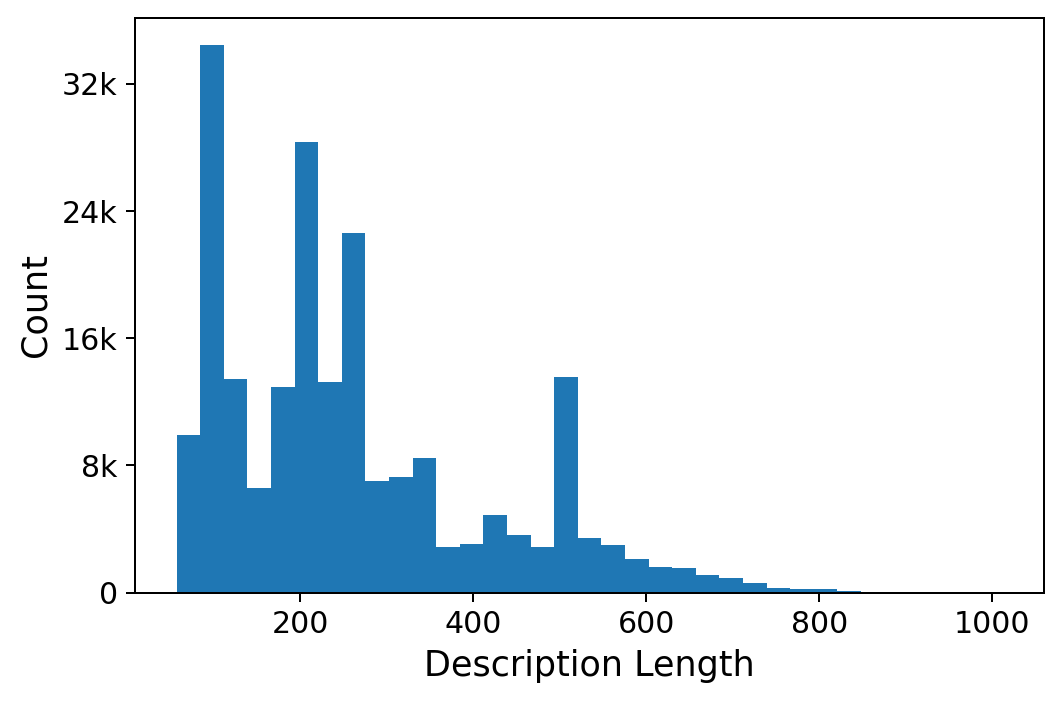}
        \caption{Description length.}
        \label{fig:hist_description_len}
    \end{subfigure}
    \hfill
    \begin{subfigure}[t]{0.25\textwidth}
        \centering
        \includegraphics[width=\linewidth]{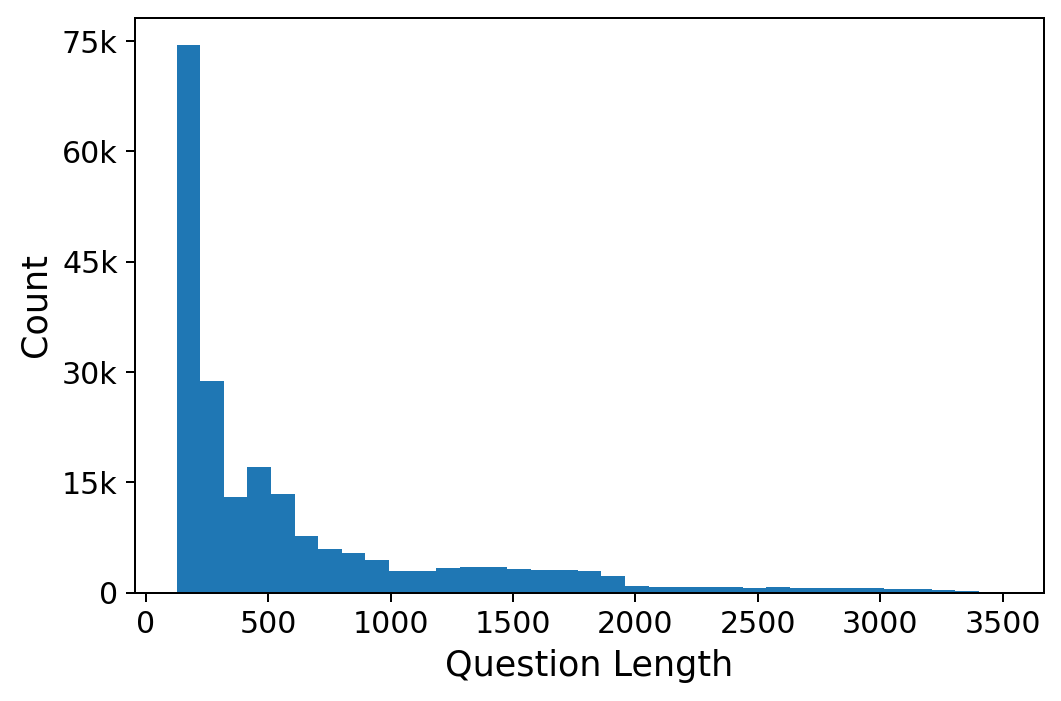}
        \caption{Question length.}
        \label{fig:hist_q_len}
    \end{subfigure}
    \hfill
    \begin{subfigure}[t]{0.20\textwidth}
        \centering
        \includegraphics[width=\linewidth]{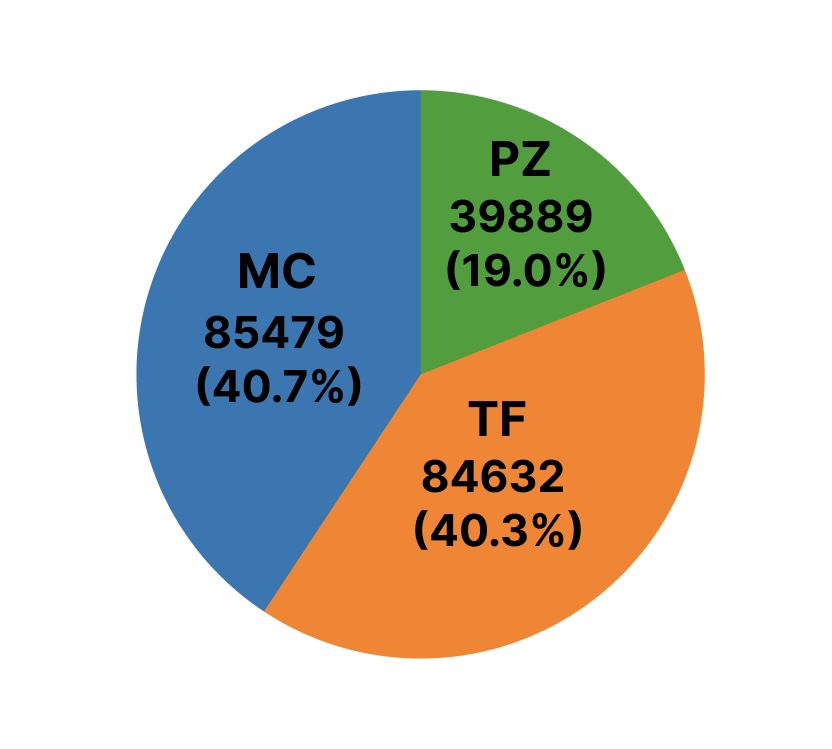}
        \caption{Question type dist.}
        \label{fig:q_type_distribution}
    \end{subfigure}
    \caption{Histograms of time series, description and question lengths, and question type distribution.}
    \label{fig:hist}
\end{figure*}

Figure~\ref{fig:domain_task_distribution} shows the distribution of domains and tasks, and Figure~\ref{fig:q_type_distribution} shows the distribution of question types. Samples are nearly balanced across tasks, question types, and major domains.
Figures~\ref{fig:hist_ts_len}–\ref{fig:hist_q_len} present the histograms of time series length, description length, and question length, all of which exhibit long-tail distributions.

\subsection{Evaluation Protocol}
\label{subsec:evaluation_protocol}
The \trqa benchmark includes 3 question types, each with a specific evaluation metric. \textit{TF} and \textit{MC} questions are evaluated using accuracy. \textit{PZ} questions are scored by comparing each predicted position with the ground truth and computing the proportion of correct matches.
For example, with a ground truth {A, B, C, D} and prediction {B, A, C, D}, only the last two match, yielding 50\% accuracy. 

\section{Experiments}
\label{sec:experiments}
In this section, we present experimental results of both commercial and open-source LLMs and language-capable TSFMs on our \trqa benchmark, and provide analysis of the results.

\subsection{Main Results}
We evaluate \emph{zero-shot} performance of (1) commercial LLMs: GPT-4.1, GPT-4o, Claude-3.5-Sonnet and Gemini-2.5-Flash;
(2) medium size open-source LLMs: Qwen3-8B~\citep{qwen3}, LLaMA3.1-8B~\citep{dubey2024llama}, Ministral-8B;
(3) small size open-source LLMs: Qwen3-0.6B~\citep{qwen3}, LLaMA3.2-1B~\citep{dubey2024llama}, Gemma3-1B~\citep{team2025gemma}.
We further apply \emph{instruction tuning}~\cite{peng2023instruction} to the open-source LLMs using LoRA~\cite{hu2022lora}.
We set LoRA rank as 16, fix learning rate as $10^{-5}$ with cosine schedule, and train models for 2 epochs on a single A100 GPU. We also evaluate language-capable TSFMs: ChatTime~\citep{wang2024chattimeunifiedmultimodaltime} and ChatTS~\citep{xie2025chattsaligningtimeseries}.

\noindent \textbf{Overall Results.}
The rightmost column in Table~\ref{tab:main_results} presents averaged results over all the samples (not simply over each row).
(1) \emph{Zero-shot}: Commercial LLMs consistently outperform open-source LLMs, and medium-sized (8B) open-source models outperform small (1B) ones.
(2) After \emph{instruction tuning}: All open-source models improve substantially; notably, Gemma3-1B (69.70) surpasses Gemini-2.5-Flash (65.08).
These results indicate that instruction tuning can markedly enhance open-source models, narrowing the performance gap with and even outperform commercial LLMs.
Additionally, results on TSFMs further demonstrate that \trqa can extend beyond LLM evaluation and serve as benchmark for any language-capable models.  

\noindent \textbf{Task-Level Results.} 
(1) \textit{Conventional Analysis.} In zero-shot settings, both commercial and open-source LLMs perform poorly on anomaly detection and classification, but open-source models improve markedly after instruction tuning (e.g., LLaMA-3.1-8B reaches 91.02 and 91.27).
(2) \textit{Advanced Analysis.} For characterization and comparison, commercial models outperform medium-sized open-source models, likely due to broader pretraining exposure. 
Data transformation and temporal relationship, especially \textit{PZ} questions, remain difficult for all models.

\noindent \textbf{Question Type-Level Results.} 
Across the three question types (\textit{TF}, \textit{MC}, \textit{PZ}), open-source models perform best on \textit{TF}, worse on \textit{MC}, and poorest on \textit{PZ}. Performance on \textit{PZ} is substantially lower than on \textit{TF} and \textit{MC}, in both zero-shot and tuned settings. Considerable room for improvement remains, e.g., the best \textit{PZ} score is only 67.68. 

\subsection{Analysis}
We use best performing commercial LLMs, i.e., Gemini-2.5-Flash and GPT-4.1, and open-source LLMs, i.e., LLaMA3.1-8B and Qwen3-8B to conduct further analysis. To examine their analytical ability on the proposed \trqa Benchmark, we cover three key perspectives:
\textit{Accuracy Correlate Analysis}, \textit{Task-Specific Analysis}, and \textit{Case Study}.

\subsubsection{Accuracy Correlate Analysis}
\label{Accuracy Correlate Analysis}
\textbf{Input Lengths.}
Figure~\ref{fig:input_length_accuracy} illustrates the relationship between input length and model accuracy. 
Across all five tasks, except the Temporal Relationship task, we observe a consistent trend that performance declines as input length increases, indicating that longer inputs correspond to more difficult questions. However, the Temporal Relationship task exhibits the opposite behavior, where accuracy improves with increasing input length. The analysis is shown in Figure ~\ref{fig:qt_input_length_accuracy}, which the newly proposed \textit{PZ} question exhibits the opposite trend. The fact that \textit{PZ} performance scales positively with length proves that models are actively utilizing global context to deduce the correct chronological order, proving that \textit{PZ}  question is a rigorous probe for \textit{Global Causal Reasoning}. More details in Appendix~\ref{appendix:accuracy_correlate_analysis}.

\noindent \textbf{Topics \& Subtopics vs. Accuracy.} Tasks such as Characterization and Comparison include questions with different selected topics and subtopics from a predefined list~\ref{table:topics}. To understand how the complexity of topics and subtopics influences model performance, we analyzed the relationship between the number of topics and subtopics used in each question and the corresponding model accuracy. Based on Table~\ref{tab:subtopic-topic-analysis}, we observe that the complexity of questions with varying number of topics and subtopics doesn't have direct impact on model accuracy, indicating that the \trqa Benchmark is largely unbiased. We further analyzed the difficulty of individual topics by examining how different topic combinations influence model performance across tasks (Figure~\ref{fig:topics123}) and found that questions with topics such as \textit{seasonality}, \textit{autocorrelation}, \textit{dispersion}, and \textit{noise} are harder to models. More details are provided in Appendix~\ref{appendix:accuracy_correlate_analysis}.

\noindent \textbf{Domain vs. Accuracy.} We conducted an in-depth analysis of how domain variation impacts overall model accuracy on \trqa Benchmark. Results are summarized in Table~\ref{tab:domain_accuracy}. Our analysis reveals that questions from domains including \textit{Synthetic}, \textit{IT}, \textit{Robotics}, and \textit{Web} pose greater challenges to models under the zero-shot setting, while questions from \textit{Sales} and \textit{Web} domains remain the most difficult after instruction tuning. More details are provided in Appendix~\ref{appendix:accuracy_correlate_analysis}.

\subsubsection{Task Specific Analysis}

\noindent \textbf{Data Transformation.}
We analyze model performance on the Data Transformation task, which is designed to evaluate a model's understanding of 3 transformation operators: Fourier Transform (FT), Wavelet Transform (WT), and First-Order Differencing (FOD). 
For each operator, we assess performance by measuring the accuracy on both \textit{MC} and \textit{TF} questions. As shown in Table~\ref{tab:analysis_datatransformation}, for zero-shot evaluation, our key finding highlights a limitation in which both commercial and open-source models fail to provide accurate answers, except for FOD. More details are provided in Appendix~\ref{appendix:task_specific_analysis}.

\noindent \textbf{Temporal Relationship.} Beyond the input length analysis in Section~\ref{Accuracy Correlate Analysis}, we further examined how domain-level information influences model performance on \textit{PZ} questions. Results are summarized in Table~\ref{tab:pz_domain_accuracy}, which \textit{Web} and \textit{Sales} domains remain the most challenging across both zero-shot and instruction-tuning settings. 
To identify the cause, we analyzed the boundary consistency of incorrect predictions and identified a significant \textit{Smoothness Bias}. As shown in Table~\ref{tab:smoothness_bias_detailed}, models consistently attempt to repair legitimate discontinuities by predicting transitions that are smoother than the ground truth. This failure highlights the critical utility of the \textit{PZ} task: since legitimate volatility varies by domain, \textit{PZ} acts as a rigorous discriminator for Temporal Fidelity. It penalizes models that rely on generic smoothing priors and rewards those that capture the irregular dynamics of the target domain. 
More details are provided in Appendix~\ref{appendix:task_specific_analysis}.

\noindent \textbf{Comparison.} For Comparison task, we investigate whether providing explicit domain-level context affects model accuracy. The task requires comparing two input time series, which we test under two conditions: (1) when both series originate from the same domain and (2) when they are from different domains. In both scenarios, the corresponding domain names are provided to the model as textual description.  As shown in Table~\ref{tab:analysis_comparison}, we observe no significant performance difference between the same-domain and different-domain settings across either \textit{MC} or \textit{TF} questions. This suggests that the Comparison Task is domain invariant. More details are provided in Appendix~\ref{appendix:task_specific_analysis}.

\subsubsection{Case Study.} 
\textbf{First Letter Distribution.} To explore potential biases in model behavior, we analyzed the distribution of the first letters in model responses for \textit{PZ} questions (Figure~\ref{fig:first_letter}). An interesting pattern emerges: for incorrectly answered questions, Qwen3-8B tends to output \emph{C} more frequently, whereas LLaMA3.1-8B tends to output \emph{A}. 

\noindent \textbf{Incorrect Output Format.} While models generally demonstrate a strong understanding of the expected response format for common question types such as \textit{MC} and \textit{TF}, we include sample responses from each model with unique or unexpected behaviors - cases where the models do not adhere to the specified instructions for the \textit{PZ} question format.  More details are provided in Appendix~\ref{appendix:case_study}.

\subsection{Human Evaluation}
\label{subsec:human_eval}
We conduct human evaluations of the multi-LLM consensus labels for characterization and comparison (Section \ref{subsec:benchmark_construction}). 
Six Ph.D.-level experts manually annotate 600 questions (300 each), serving as ground truth. Uncertain or problematic QA pairs are flagged, multiple answers allowed when valid, and explanations provided for disagreements with the benchmark. 

Our evaluation yields two main findings.
(1) Question quality: Uncertainty rates are low (5\% for characterization, 7\% for comparison), showing that most questions are clear.
(2) Answer accuracy: For unambiguous cases, benchmark answers align with human judgments in 91.2\% of characterization and 87.4\% of comparison.
These results indicate that the automatic pipeline produces reliable QA pairs, though comparison remains harder, with lower agreement and higher uncertainty (Figure~\ref{fig: mismatch}). Details in Appendix~\ref{appendix:human_evaluation}.

\section{Conclusion}
\trqa establishes a large-scale benchmark for time series question answering, comprising 210k samples curated from 13 domains and covering 6 tasks with 3 question types. It provides a unified platform for probing the strengths and limitations of language-capable models on time series analysis.

Our results show that, despite progress from instruction tuning, substantial challenges remain, particularly for advanced analysis and puzzling questions, highlighting significant room for improvement in current models. By intentionally focusing on standardized closed-ended analytical reasoning with reproducible evaluation protocols, \trqa establishes a benchmark foundation upon which future work can build. In this way, \trqa may serve not only as an evaluation benchmark, but also as a training resource for future language-capable time-series reasoning models, including extensions toward multivariate settings, multimodal inputs, open-ended analysis, and agentic systems.

\section*{Limitations}
While \trqa provides a unified and large-scale benchmark for time series question answering, several limitations remain. First, although the benchmark spans 13 domains and 6 analytical tasks, it does not fully capture the diversity of real-world temporal processes. Many application settings involve irregular sampling, strong exogenous drivers, or domain-specific structures, which are only partially reflected in our datasets. Expanding \trqa toward irregular, mixed-frequency, and exogenous-aware scenarios would further improve realism. Second, our task taxonomy primarily focuses on analytical capabilities that can be expressed through structured questions. However, real systems often require richer forms of temporal reasoning, suggesting opportunities to design tasks that more directly probe these behaviors. Third, while the newly proposed puzzling question encourages global structural reasoning, it may penalize models biased toward locally smooth transitions and introduces higher computational costs. Future extensions could incorporate complementary formats that disentangle local continuity, long-range consistency, and domain-specific volatility. Finally, although \trqa covers multiple domains, the benchmark remains static, whereas real deployments face evolving distributions and emerging domains. Building dynamic extensions that evaluate adaptation and robustness under distribution shifts represents an important next step.

\section*{Ethical Considerations}
All datasets and language models used in this work are publicly available. The \trqa dataset was constructed from established, publicly accessible time series benchmarks and synthetic data generation followed ethical guidelines to minimize biases and ensure data quality.

\section*{Acknowledgments}
We leverage Large Language Models (LLMs) from two perspectives:
(1) Polishing the writing, where LLMs are used to refine the clarity, fluency, and consistency of the paper; and (2) Labeling, where LLMs assist in generating high-quality question-answer (QA) pairs and providing preliminary annotations, which are then validated or aggregated through consensus to create reliable ground-truth labels.

\bibliography{custom}

\appendix
\newpage
\section{Data Collection}
\label{appendix:data_collection}
In this section, we detail the data sources, including \emph{core datasets} (Appendix~\ref{appendix:core_dataset}), \emph{anomaly detection datasets} (Appendix~\ref{appendix:anomaly_detection_dataset}), and \emph{classification datasets} (Appendix~\ref{appendix:classification_dataset}).

\subsection{Core Datasets} 
\label{appendix:core_dataset}

\begin{table}[h]
\centering
\caption{Summary of the core datasets.}
\label{tab:core_data_appendix}
\resizebox{0.95\linewidth}{!}{%
    \begin{tabular}{lcc}
    \toprule
    \textbf{dataset\_name} & \textbf{total\_data\_point} & \textbf{domain} \\
    \midrule
    AustralianElectricityDemand & 1,153,584 & energy \\
    BDG-2 Rat & 4,728,288 & energy \\
    GEF12 & 788,280 & energy \\
    ExchangeRate & 56,096 & finance \\
    FRED MD & 76,612 & finance \\
    BIDMC32HR & 8,000,000 & healthcare \\
    PigArtPressure & 624,000 & healthcare \\
    USBirths & 7,275 & healthcare \\
    Sunspot & 73,924 & nature \\
    Saugeenday & 23,711 & nature \\
    SubseasonalPrecip & 9,760,426 & nature \\
    HierarchicalSales & 212,164 & sales \\
    m5 & 58,327,370 & sales \\
    PedestrianCounts & 3,130,762 & transport \\
    PEMS03 & 9,382,464 & transport \\
    UberTLCHourly & 1,129,444 & transport \\
    WikiDaily100k & 274,099,872 & web \\
    \bottomrule
    \end{tabular}%
}
\end{table}

We extract data from multiple time-series datasets including: Australian Electricity Demand~\citep{godahewa2021monash}, BDG-2 Rat~\citep{bdg2_2020}, GEF12~\citep{hong2014gefcom2012}, ExchangeRate\citep{lai2018lstnet}, FRED MD\citep{mccracken2016fredmd}, BIDMC32HR~\citep{tan2020tser}, PigArtPressure~\citep{dau2019ucr}, USBirths~\citep{godahewa2021monash}, Sunspot~\citep{godahewa2021monash}, Saugeenday~\citep{godahewa2021monash}, SubseasonalPrecip~\citep{mouatadid2024subseasonal}, HierarchicalSales~\citep{mancuso2021eswa}, M5~\citep{makridakis2022m5}, PedestrianCounts~\citep{cityofmelbourne_pedcounts}, PEMS03~\citep{caltrans_pems}, UberTLCHourly~\citep{fivethirtyeight2015uber},WikiDaily100k~\citep{ansari2024chronos}. Below are some more detailed descriptions on those datasets. 

\noindent \textbf{Australian Electricity Demand.} A single long time series from the Monash Time Series Archive representing half-hourly electricity demand for Victoria, Australia in 2014 (17{,}520 observations), extracted from the R package \texttt{fpp2} (dataset name: ``elecdemand''). Temperatures corresponding to each demand value are available in the original dataset.

\noindent \textbf{BDG-2 Rat.} From The Building Data Genome Project 2 (MIT License), consisting of measurements from 3{,}053 meters across 1{,}636 commercial buildings over 2016--2017. One or more meters per building measured total electrical, heating and cooling water, steam, solar energy, water, and irrigation usage. We use the whole-building electricity meter measurements from the Bear, Fox, Panther, and Rat sites, totaling 611 buildings (from the CSV file \texttt{electricity\_cleaned.csv}).

\noindent \textbf{GEF12.} A benchmark compiled from the Global Energy Forecasting Competition 2012 (load forecasting tracks), containing 20 aggregated-level hourly load series and 11 temperature series from 2004-01-01 00:00 to 2008-06-30 05:00. Because the one-to-one correspondence between temperature and load series is not clearly defined, a common strategy is to use a single temperature series for all loads (here, the second temperature series). The dataset is competition-grade and was used without additional preprocessing; visualizations show obvious periodicity and seasonality in the aggregated loads.

\noindent \textbf{ExchangeRate.} Daily exchange rates for currencies of eight countries---Australia, United Kingdom, Canada, Switzerland, China, Japan, New Zealand, and Singapore---covering 1990 to 2016.

\noindent \textbf{FRED-MD.} 107 monthly time series of macro-economic indicators from the Federal Reserve Bank, starting from 1959-01-01, extracted from the FRED-MD database.

\noindent \textbf{BIDMC32HR.} Derived from BIDMC ICU recordings: PPG and respiratory signals/IP (sampling rate 125~Hz) from 53 adult patients, with breath annotations used to form reference targets in the source dataset. Following the adaptation in subsequent work, PPG and ECG are converted into 32-second sliding-window time series; the average heart rate (HR) in each 32~s window is the target. The datasets are split by randomly selecting 30\% as test, yielding 5{,}550 training and 2{,}399 test time series.

\noindent \textbf{PigArtPressure.} Based on a source dataset from 52 pigs with three vital signs monitored before and after an induced injury. Three datasets are created: AirwayPressure (airway pressure), ArtPressure (arterial blood pressure), and CVP (central venous pressure).

\noindent \textbf{US Births.} A single long daily time series of the number of births in the United States from 1969-01-01 to 1988-12-31 (7{,}305 observations), extracted from the R package \texttt{mosaicData}.

\noindent \textbf{Sunspot.} A single long daily time series of sunspot numbers from 1818-01-01 onward, with additional related series (monthly means, smoothed series, yearly totals, hemispheric series) in the original source. The repository used here contains the daily series from 1818-08-01 to 2020-05-31 and includes both the raw data (with missing values) and an LOCF-imputed version.

\noindent \textbf{Saugeen.} A single long daily time series of the Saugeen River mean flow at Walkerton (in cubic meters per second) from 1915-01-01 to 1979-12-31 (23{,}741 observations), extracted from the R package \texttt{deseasonalize} (dataset name: ``SaugeenDay'').

\noindent \textbf{Subseasonal Precipitation.} Extracted from SubseasonalClimateUSA: daily precipitation measurements (millimeters) for a single $1.5^\circ \times 1.5^\circ$ latitude--longitude grid cell, covering 1948--1978.

\noindent \textbf{Hierarchical Sales.} 118 daily time series of SKU-level sales for four national pasta brands from 2014-01-01 to 2018-12-31, including a binary indicator for promotion. The series can be organized into a three-level hierarchy.

\noindent \textbf{M5.} The M5 ``Accuracy'' competition dataset requiring point forecasts for 30{,}490 bottom-level daily series that aggregate to 42{,}840 time series representing hierarchical unit sales for Walmart. The competition paper details the implementation, results, top methods, and implications for forecasting research.

\noindent \textbf{Pedestrian Counts.} Hourly pedestrian counts from 66 sensors in Melbourne starting from May 2009. The original data are updated monthly; the repository snapshot used here contains counts up to 2020-04-30.

\noindent \textbf{PEMS03.} Datasets sourced from Caltrans PeMS, which collects 30-second traffic readings and aggregates them into 5-minute intervals (288 time steps per day). Road network structure is derived from connectivity status and actual distances between sensors.

\noindent \textbf{Uber TLC Daily.} Counts of Uber pick-ups from various New York City locations between January and June 2015, obtained from FiveThirtyEight's ``uber-tlc-foil-response'' repository and aggregated at hourly and daily resolutions.

\noindent \textbf{WikiDaily10k.} Daily traffic data for 10{,}000 Wikipedia pages.

\subsection{Anomaly Detection Dataset}
\label{appendix:anomaly_detection_dataset}
We extract data from multiple time-series anomaly detection benchmarks~\cite{paparrizos2022tsb, DBLP:conf/kdd/SuZNLSP19}, including ECG~\citep{moody2001impact}, SMD~\citep{DBLP:conf/kdd/SuZNLSP19}, MGAB~\citep{thill10markusthill} Genesis~\citep{von2018anomaly}, GHL~\citep{DBLP:journals/corr/FilonovLV16}, Occupancy~\citep{candanedo2016accurate}.
These datasets span various domains, including healthcare (ECG), mathematical biology (MGAB), spacecraft telemetry (Genesis), industrial control system (GHL), environmental sensing (Occupancy), cyber-security on IT Operations (SMD). The statistics of these datasets are shown in Table~\ref{tab:dataset_stats_anomaly}. To address class imbalance, we count the number of anomalous sequences and randomly select an equal number of normal sequences, resulting in a balanced dataset. 
Below are the meta information for each dataset.
\begin{table}[h]
\centering
\resizebox{0.95\linewidth}{!}{%
    \begin{tabular}{lcccc}
    \toprule
    \textbf{Name} & \textbf{\# Samples} & \textbf{Domain} \\
    \midrule
    ECG         & 17,862 & Healthcare \\
    SMD         & 58,888 & Cyber-security on IT Operations \\
    MGAB        & 376 & mathematical biology \\
    Genesis     & 274 & Spacecraft Telemetry \\
    GHL         & 768 & Industrial Control System \\
    Occupancy   & 8,178 & Environmental Sensing \\
    \bottomrule
    \end{tabular}%
}
\caption{Summary of anomaly detection datasets.}
\label{tab:dataset_stats_anomaly}
\end{table}

\noindent \textbf{MGAB.} This dataset is composed of Mackey-Glass time series with non-trivial anomalies. Mackey-Glass time series exhibit chaotic behavior that is difficult for the human eye to distinguish.

\noindent \textbf{ECG.} This dataset is a standard electrocardiogram dataset and the anomalies represent ventricular premature contractions. The ECG recordings were made using Del Mar Avionics model 445 two-channel reel-to-reel Holter recorders, and the analog signals were recreated for digitization using a Del Mar Avionics model 660 playback unit. The digitization rate (360 samples per second per channel) was chosen to accommodate the use of simple digital notch filters to remove 60 Hz (mains frequency) interference.

\noindent \textbf{Genesis.} This dataset is a portable pick-and-place demonstrator which uses an air tank to supply all the gripping and storage units. Data samples were taken through an OPC connection with a resolution of 50 milliseconds for a total of 42 production cycles. The first 38 production cycles contain only normal behavior and were used to train the selforganizing map for both experiments shown in this section. Two of the 4 remaining cycles contain anomalous behavior and are used for the anomaly detection.

\noindent \textbf{GHL.} This dataset is a Gasoil Heating Loop Dataset and contains the status of 3 reservoirs such as the temperature and level. Anomalies indicate changes in max temperature or pump frequency. Type of cyber attack to the normal process logic is the unauthorized change of max Receiving Tank level. By changing the time of attack and the value of the hacked max Receiving Tank level, we generated many anomalous data sets used for fault detection.

\noindent \textbf{Occupancy.} This dataset contains experimental data of room occupancy, such as temperature, humidity, light, and CO2. Ground-truth occupancy was obtained from time stamped pictures that were taken every minute.

\noindent \textbf{SMD.} SMD (Server Machine Dataset) is collected from a large Internet company. The data is sampled every 5 seconds. Labels denote whether a point is an anomaly and the dimensions contribute to every anomaly.

\subsection{Classification Dataset}
\label{appendix:classification_dataset}
We extract data from the UCR Archive~\citep{dau2019ucr}. To create a focused subset for our study, we applied two primary selection criteria: we included only datasets with four or fewer classes and time series
with a sequence length of 400 time points or less. Through our selection, we extract data from 37 benchmarks in the UCR Archive, including SonyAIBORobotSurface1 \& SonyAIBORobotSurface2~\citep{10.1145/2020408.2020587}, FreezerRegularTrain \& FreezerSmallTrain~\citep{freezer}, ToeSegmentation1 \& ToeSegmentation2~\citep{twosegmentation}, TwoPatterns~\citep{twopattern}, CBF~\citep{Saito1994LocalFE}, Wafer \& ECG200~\citep{wafer}, TwoLeadECG, ECGFiveDays, DistalPhalanxOutlineCorrect \& MiddlePhalanxOutlineCorrect \& ProximalPhalanxOutlineCorrect \& DistalPhalanxOutlineAgeGroup \& MiddlePhalanxOutlineAgeGroup \& ProximalPhalanxOutlineAgeGroup \&  PhalangesOutlinesCorrect~\citep{DBLP:journals/corr/BagnallD14}, MoteStrain~\citep{motestrain}, GunPointMaleVersusFemale \& GunPointOldVersusYoung \& GunPointAgeSpan \& GunPoint~\citep{gunpoint}, Strawberry~\citep{strawberry}, ItalyPowerDemand~\citep{4053126}, Chinatown, BME, PowerCons, DodgersLoopWeekend \& DodgersLoopGame~\citep{10.1145/1150402.1150428}, DiatomSizeReduction, SmoothSubspace~\citep{10.1016/j.ins.2016.05.040}, UMD, Wine, Coffee~\citep{briandet:hal-01606904}, and ArrowHead~\citep{10.1145/1557019.1557122}. These datasets span various domains, including robotics, energy, healthcare, synthetic, manufacturing, nature, and transport. The statistics of these datasets are shown in Table~\ref{tab:dataset_classification}.

\begin{table}[h]
\centering
\resizebox{0.99\linewidth}{!}{%
    \begin{tabular}{lcccc}
    \hline
    \textbf{Name} & \textbf{\# Samples} & \textbf{\# Classes} & \textbf{Domain} \\
    \hline
    SonyAIBORobotSurface1  & 486 & 2 & Robotics \\
    SonyAIBORobotSurface2  & 771 & 2 & Robotics \\
    FreezerRegularTrain    & 2,404 & 2 & Energy \\
    FreezerSmallTrain      & 2,353 & 2 & Energy \\
    ToeSegmentation1       & 210 & 2 & Healthcare \\
    ToeSegmentation2       & 129 & 2 & Healthcare \\
    TwoPatterns            & 3,999 & 4 & Synthetic \\
    CBF                    & 757 & 3 & Synthetic \\
    Wafer                  & 5,744 & 2 & Manufacturing \\
    ECG200                 & 159 & 2 & Healthcare \\
    TwoLeadECG             & 923 & 2 & Healthcare \\
    ECGFiveDays            & 704 & 2 & Healthcare \\
    DistalPhalanxOutlineCorrect      & 690 & 2 & Healthcare \\
    MiddlePhalanxOutlineCorrect      & 731 & 2 & Healthcare \\
    ProximalPhalanxOutlineCorrect    & 688 & 2 & Healthcare \\
    DistalPhalanxOutlineAgeGroup     & 423 & 3 & Healthcare \\
    MiddlePhalanxOutlineAgeGroup     & 435 & 3 & Healthcare \\
    ProximalPhalanxOutlineAgeGroup   & 485 & 3 & Healthcare \\
    PhalangesOutlinesCorrect         & 2,076 & 2 & Healthcare \\
    MoteStrain                       & 1,012 & 2 & Nature \\
    GunPointMaleVersusFemale         & 362 & 2 & Healthcare \\
    GunPointOldVersusYoung           & 356 & 2 & Healthcare \\
    GunPointAgeSpan                  & 368 & 2 & Healthcare \\
    GunPoint                         & 169 & 2 & Healthcare \\
    Strawberry                       & 786 & 2 & Nature \\
    ItalyPowerDemand                 & 890 & 2 & Energy \\
    Chinatown                        & 293 & 2 & Transport \\
    BME                              & 137 & 3 & Synthetic \\
    PowerCons                        & 294 & 2 & Energy \\
    DodgersLoopWeekend               & 111 & 2 & Transport \\
    DodgersLoopGame                  & 115 & 2 & Transport \\
    DiatomSizeReduction              & 248 & 4 & Nature \\
    SmoothSubspace                   & 236 & 3 & Synthetic \\
    UMD                              & 148 & 3 & Synthetic \\
    Wine                             & 85 & 2 & Nature \\
    Coffee                           & 48 & 2 & Nature \\
    ArrowHead                        & 175 & 3 & Nature \\
    \hline
    \end{tabular}%
}
\caption{Classification data used in our experiments.}
\label{tab:dataset_classification}
\end{table}

\newpage
\section{Benchmark Construction}
\label{appendix:benchmark_construction}
In this section, we provide extra content about the construction process for each task and provide examples of each task.

\subsection{Hierarchical Uniform Sampling}
    \begin{algorithm}[h]
    \small
    \SetAlCapFnt{\small}
    \caption{Hierarchical Random Sampling}
    \label{algo:hierarchical_random_sampling}
    \KwInput{
    Domains $\mathcal{M}$;\\ 
    \quad Datasets $\mathcal{D}(m)$ for each domain $m\in\mathcal{M}$;\\ 
    \quad Sequences $\mathcal{S}(d)$ for each dataset $d\in\mathcal{D}$;\\ 
    \quad Segment length $l$}
    \KwOutput{Segment $s_{t:t+l-1}$}
    \BlankLine
    $m \gets \text{UniformPick}(\mathcal{M})$\tcp*{Randomly select a domain}
    $d \gets \text{UniformPick}(\mathcal{D}(m))$\tcp*{Randomly select a dataset in the domain}
    $s \gets \text{UniformPick}(\mathcal{S}(d))$\tcp*{Randomly select a seq. from the dataset}
    $t \gets \text{UniformPick}\{1,\dots,|s|-l+1\}$\tcp*{Randomly select a start index}
    \Return{$s_{t:t+l-1}$}\tcp*{Return the segment}
\end{algorithm}
For all the advanced reasoning tasks, including characterization, comparison, data transformation and temporal relationship, all the input time series are sampled from the \emph{core dataset} (Appendix~\ref{appendix:core_dataset}).
To ensure a balanced distribution over domains, datasets and sequences, we use \emph{Hierarchical Uniform Sampling} presented in Algorithm~\ref{algo:hierarchical_random_sampling} to obtain samples.

\subsection{Data Bias}
\label{appendix:data-bias}
Unless otherwise specified, all samples have a random length in $[32, 256]$, and are z-scored to reduce data bias. The term data bias refers specifically to scale-based shortcuts or magnitude variance across heterogeneous domains, rather than semantic or sampling bias. We justify the use of z-score normalization on 2 main grounds: (1) \textit{Preventing Magnitude-Based Shortcuts}, (2) \textit{Standard Practice and Task Alignment}. 

\textit{Preventing Magnitude-Based Shortcuts}: \trqa is a unified benchmark that aggregates data from 13 distinct domains, each possessed of vastly different magnitudes and units. Without normalization, large language models (LLMs) could exploit these scale differences as shortcuts to identify the source domain or dataset without performing genuine temporal reasoning. Normalization prevents this risk, forcing the model to rely on structural reasoning rather than memorizing absolute value ranges. \\

\begin{table*}[t]
\centering
\small
\begin{tabular}{c|p{10cm}}
\toprule
\textbf{Topic} & \textbf{Sub-Topics} \\ 
\midrule
Trend & trend directions, trend types, trend shapes, trend strength, structural breaks, global and local trends \\ 
\midrule
Seasonality & seasonality period, seasonal strength, multiple seasonality patterns, changing seasonality \\ 
\midrule
Cyclicity & amplitude, peaks and trough, duration \\ 
\midrule
Noise & noise level, global and local noise \\ 
\midrule
Stationarity & stationarity strength, global and local stationarity, types of non-stationarity \\ 
\midrule
Autocorrelation & types of autocorrelation, autocorrelation structures, lags, mean-reversion, persistence of autocorrelation \\ 
\midrule
Dispersion & basic measures of variability (variance level), relative measures (signal-to-noise ratio level), coefficient of variation level, time-varying dispersion (volatility, heteroskedasticity), entropy, multi-scale dispersion \\ 
\midrule
Shape & global shapes, local shapes, shapelets, motifs, curves, change points, pattern complexity \\ 
\midrule
Irregularity & mean shift, variance shift, trend shift, seasonality irregularity, cyclic shift, distributional change, structural breaks, autocorrelation change \\ 
\midrule
\makecell{Correlation\\ (Comparison only)} & causal relationship, correlation strength, correlation types, correlation direction, cross-correlation, time-varying correlation (rolling correlation), lagged correlation, global and local correlations, correlation of decomposed components \\ 
\bottomrule
\end{tabular}
\caption{Topics and Sub-Topics for Time Series Analysis}
\label{table:topics}
\end{table*}

\textit{Standard Practice and Task Alignment}: While real-world data is indeed not standardized, normalization is a ubiquitous and necessary preprocessing step in the time series literature to ensure numerical stability and cross-domain comparability. This approach aligns with established protocols in widely used benchmarks such as the UCR Archive~\citep{dau2019ucrtimeseriesarchive}, and recent time series foundation model studies like Time-LLM~\citep{jin2024timellmtimeseriesforecasting} and Chronos~\citep{ansari2024chronoslearninglanguagetime}, which consistently utilize normalization or scaling to handle distribution shifts. Additionally, the core objective of \trqa is to evaluate reasoning capabilities. Z-score normalization is a linear transformation that preserves the fundamental properties required for these tasks while removing the confounding factor of arbitrary absolute magnitudes.

\subsection{Characterization}
\label{appendix:characterization}

The characterization task assesses the model’s capability to analyze fundamental properties of time series, including trend, seasonality, and dispersion. Questions are posed as \textit{TF} or \textit{MC}, and final answers are determined through multi-LLM consensus.

Each instance consists of a univariate time series sample $\mathbf{x}$ with associated metadata (text description, domain, dataset). 
Given a sample $\mathbf{x}$ and its metadata, we instruct GPT-4o~\citep{hurst2024gpt} to generate one QA pair per instance using a randomly selected subset of one to three topics (from Table~\ref{table:topics}) and a question type (TF or MC). The process is as follows.

\textit{Step 1: Captioning \& sub-topic selection.}
GPT first produces a short, neutral caption summarizing visible patterns (e.g., “gradual upward drift with weak weekly oscillation”). For each chosen topic, a sub-topic is sampled uniformly at random, e.g., trend, seasonality and dispersion.

\textit{Step 2: QA synthesis.}
GPT generates a TF or MC question grounded in $\mathbf{x}$, the caption, and the selected sub-topics.

\textit{Step 3: Self-verification.}
GPT performs a self-check and outputs a confidence score in [0,1]. 
We retain QA pairs only if confidence $\ge$ 0.95.

\textit{Step 4: Multi-LLM consensus.}
We query GPT-4.1, Gemini-2.5-Flash, and Claude-3.5-Sonnet using the same prompt, which includes the generated question along with its allowed answer choices (for both TF and MC formats), and collect their responses.
To determine the final label, we adopt a weighted majority voting scheme among these three models and GPT-4o’s original answer.
Specifically, GPT-4.1 and Gemini-2.5-Flash are assigned higher weights of 1.5 each, reflecting their superior performance in preliminary evaluations, while Claude-3.5-Sonnet and GPT-4o are each assigned a weight of 1.0.
The option with the highest total weighted vote is selected as the consensus answer.
If a tie occurs—i.e., two or more answers receive the same highest weighted score—the corresponding QA pair is discarded to avoid introducing ambiguity or noise into the dataset.
This ensemble-based strategy mitigates single-model biases, smooths out random errors, and produces more reliable and stable labels, which are crucial for ensuring the benchmark’s quality.

\noindent Here's the \emph{system} prompt template.
\begin{lstlisting}[style=promptstyle]
You are an expert of time series analysis.
1. Generate a meta_caption solely based on the meta information within 50 words. 
2. Generate a detailed_caption based on both meta information and time series within 100 words.
3. Generate a {} based on the time series, meta_caption, detailed_caption and the more detailed question instructions.
4. Generate a correct answer {} for your question.
5. A successful generation must meet the following conditions:
(1) there is only one correct answer;
(2) the question stricktly follows the instructions;
(3) the answer of the question cannot be easily derived from the meta_caption;
(4) the question should be about the time series itself without invovling external knowledge;
(5) do not repeat the input time series in questions or answers.
6. Show your confidence of your determination of success within 0-1.
\end{lstlisting}

\noindent Here's the \emph{user} prompt template.
\begin{lstlisting}[style=promptstyle]
The time series is {}.
Its meta information is {}.
The question must be about all these topics: {}.
The sub-topics of {} includes but not limited to {}.
First think about the all possible sub-topics and their taxonomy.
Then randomly pick a sub-topic from each topic ({}) to generate the question and answer pairs.
\end{lstlisting}

\subsection{Comparison}
\label{appendix:comparison}

The comparison task assesses the model’s ability to analyze the relative characteristics of two time series, such as overall shape, temporal alignment, and correlation patterns.
Similar to the characterization task, this task is also formulated as either \textit{TF} or \textit{MC} questions, where the model must identify similarities or differences between the given pair of sequences. The characteristics evaluated in the task are directly drawn from the standardized taxonomy of Topics and Sub-topics (from Table~\ref{table:topics}), which is shared with the Characterization task. 

To construct the comparison set, we first obtain an anchor sample $\mathbf{x}$ from a specific domain $M$, dataset $D$, and sequence $S$.
Given this anchor $\mathbf{x}$, we generate a set of ten comparison samples ${\mathbf{x}’1,\dots,\mathbf{x}’{10}}$, each having the same length as $\mathbf{x}$.
These samples are drawn in a structured manner to represent varying degrees of similarity: one from the same sequence $S$, two from different sequences within the same dataset $D$, three from other datasets within the same domain $M$, and four from entirely different domains.
This tiered sampling strategy creates a natural hierarchy of difficulty, challenging the model to distinguish between subtle intra-sequence similarities and broader cross-domain differences.

Finally, we apply a process similar to the characterization task to generate QA pairs, where GPT-based models produce questions and candidate answers.
The questions are then refined and validated through multi-LLM consensus to ensure accuracy and reduce bias, resulting in high-quality, reliable evaluation data for this task.

\subsection{Data Transformation}
\label{appendix:data_transformation}

The data transformation task evaluates the model’s ability to infer and analyze the transformation relationship between an input time series and its transformed counterpart.
These transformations are generated using well-established signal processing techniques, including the Fourier transform, wavelet transform, and first-order differencing, which are widely used in time series analysis to reveal underlying structures or remove trends.
This task is particularly challenging because it requires the model to not only recognize the patterns in the raw input series but also to understand how specific mathematical operations alter these patterns.

We use predefined templates to formulate the task as either \textit{TF} or \textit{MC} questions.
For \textit{TF} questions, the model is asked to determine whether a given candidate sequence is indeed the correct transformation of the input time series $\mathbf{x}$ (e.g., whether it is the Fourier transform result of $\mathbf{x}$).
For \textit{MC} questions, the model must select the correct transformed sequence from multiple candidates, given both the input series $\mathbf{x}$ and the specified transformation operation (e.g., Fourier transform).

To ensure accuracy and consistency, all transformations are computed using professional and reliable scientific libraries~\citep{harris2020array,virtanen2020scipy}.
The correct transformation is generated directly from the input $\mathbf{x}$, while distractor sequences are created by applying the same transformation to randomly sampled, unrelated time series $\mathbf{x}’$.
This setup forces the model to carefully analyze the relationship between the input and its transformation rather than relying on superficial similarities, providing a robust evaluation of its reasoning ability.

\noindent Here's the template to construct question.
\begin{lstlisting}[style=promptstyle]
The time series is {}.
Its meta information is {}.
The question must be about all these topics: {}.
The sub-topics of {} includes but not limited to {}.
First think about the all possible sub-topics and their taxonomy.
Then randomly pick a sub-topic from each topic ({}) to generate the question and answer pairs.
\end{lstlisting}

\subsection{Temporal Relationship}
\label{appendix:temporal_relationship}
 The Temporal Relationship task is a discriminative sequence-level reasoning task, rather than a generative forecasting task. The task evaluates a model’s ability to infer and analyze the temporal structure among sequential patches of a time series. Specifically, the task evaluates whether a model can understand the structural continuity and chronological dependencies of time series patches, testing 3 core capabilities: \textit{Structural Continuity}, \textit{Chronological Reasoning}, and \textit{Contextual Discrimination}. (1) \textit{Structural Continuity} tests whether the model can identify which candidate segment shares the underlying temporal dynamics required to validly continue a given trajectory. (2) \textit{Chronological Reasoning} tests whether the model can reconstruct the correct temporal order of shuffled patches. (3) \textit{Contextual Discrimination} tests the model's ability to distinguish the true continuation from "plausible" but incorrect alternatives that may share similar global statistics but lack local continuity.
This task is formulated as \textit{true-or-false (TF)}, \textit{multiple-choice (MC)}, or \textit{puzzling (PZ)} questions.

Given the first chronological patch $\mathbf{x}$:
(1) A TF question asks the model to determine whether a candidate patch $\mathbf{y}$ is the immediate successor of $\mathbf{x}$.
(2) An MC question requires the model to select the correct next patch from four candidates $[\mathbf{y}_1, \mathbf{y}_2, \mathbf{y}_3, \mathbf{y}_4]$.

The false candidates in both TF and MC settings are randomly sampled from the full dataset but are guaranteed to come from sequences different from that of $\mathbf{x}$, preventing the model from simply memorizing patterns.
For PZ questions, the model is presented with four shuffled successor patches of $\mathbf{x}$ and must reconstruct their correct chronological order, which poses a greater challenge as it requires deeper temporal reasoning.
All questions are generated using predefined templates to ensure consistency and diversity.

We use the following question template to construct questions.
\begin{lstlisting}[style=promptstyle]
Which of the following choices is most likely the future continuation of the given time series?
Respond ONLY with the letter of the correct choice (A, B, C, or D)

Choices:
A: {}
B: {}
C: {}
D: {}
\end{lstlisting}

\begin{lstlisting}[style=promptstyle]
Is the following patch the future continuation of the given time series?
{}
Respond ONLY with the letter of the correct choice (T or F).

Choices:
T: True.
F: False.
\end{lstlisting}

\subsection{Anomaly Detection}
\label{appendix:anomaly_detection}
First, all time-series data are standardized using z-score normalization to remove scale effects across different features. 
Next, we randomly sample a subsequence of length $T$, where $T\in[32,256]$, from each time-series instance to capture varying temporal dynamics. 
To address class imbalance, we count the number of anomalous sequences and randomly select an equal number of normal sequences, resulting in a balanced dataset. 
Finally, we enrich each sample with meta information, domain information, the normalized time-series subsequence, and its corresponding label.
        
\noindent Here's the question template.
\begin{lstlisting}[style=promptstyle]
Determine whether the given time series contains anomalies.
Respond ONLY with the letter of the correct choice (T or F).

Choices:
T: True.
F: False.
\end{lstlisting}

\subsection{Classification}
\label{appendix:classification}
Information about the time series and the task is given in the text description. 
Here's the template to construct questions.
\begin{lstlisting}[style=promptstyle]
Classify the given time series into one of the categories below.
Respond ONLY with the letter of the correct choice (A, B).

Choices:
A: {}
B: {}
\end{lstlisting}

\section{Benchmark Design Rationale}

\subsection{Benchmark Design and Scope}

Recent concurrent benchmarks have expanded time series question answering (TSQA) along several important and complementary directions. For example, SciTS~\citep{wu2026scitsscientifictimeseries} emphasizes scientific time series with heterogeneous signals and broader generation-oriented tasks, while TSRBench~\citep{yu2026tsrbenchcomprehensivemultitaskmultimodal} and MMTS-Bench~\citep{yin2026mmtsbench} extend evaluation to multimodal settings that combine textual and visual inputs. TemporalBench~\citep{weng2026temporalbenchbenchmarkevaluatingllmbased} further explores contextual and event-informed tasks for LLM-based agents. These works expand the scope of TSQA research, but also highlight that the benchmark landscape is becoming increasingly diverse in task formulation, modality, and evaluation setup.

Real-world time series applications are complex, and univariate time series analysis remains far from solved. Within this evolving landscape, \trqa spans a spectrum from fundamental analytical properties to more complex structural and relational reasoning, focusing on univariate time series and closed-ended question formats with the emphasis of scale, reproducibility, and consistent evaluation. With 210k samples, 3 question formats, and training-ready data splits, \trqa is useful not only to benchmark model performance but also to support instruction tuning and controlled comparison between model families for future extensions of TSQA research. 

\subsection{Input Representation}

In \trqa, time series inputs are represented as raw comma-separated floating-point strings after preprocessing. In particular, timestamps, trend descriptors, periodicity labels, or other analytical properties are not explicitly injected as textual features. Instead, models are expected to infer such properties directly from the numerical string representation together with the task instruction and the accompanying metadata or description. This design is consistent with \trqa’s goal of evaluating whether language-capable models can analyze fundamental time series properties from a unified natural-language QA interface.

While we acknowledge that investigating how time series values can be effectively represented within LLM token spaces remains an important open problem, we emphasize that the primary contribution of \trqa is not to propose a new encoding strategy for numerical sequences, but to consolidate time series data and analytical questions into a standardized benchmark with unified evaluation protocols. Future works may explore improved numerical or sequence encoding methodologies on top of the benchmark setting provided by TSAQA.

\subsection{Generalizability Beyond LLMs}

\trqa is a time series question answering (QA) benchmark, where models are required to consume textual prompts and descriptions together with time series inputs, and to produce answers in language-constrained formats. Directly incorporating purely numerical models would therefore require reformulating the benchmark into a different task interface. 
To demonstrate that \trqa extends beyond general-purpose LLMs, we additionally evaluate two language-capable time series foundation models (TSFMs), ChatTime~\citep{wang2024chattimeunifiedmultimodaltime} and ChatTS~\citep{xie2025chattsaligningtimeseries}, in the main results (Table \ref{tab:main_results}). Unlike conventional numerical time series models that are designed only for prediction or classification, these TSFMs can directly interact with the benchmark through natural-language QA inputs and outputs. Their inclusion therefore provides evidence that \trqa is applicable not only to general-purpose LLMs, but also to a broader class of language-capable models for time series analysis.

\newpage
\section{Examples}
\label{appendix:examples}
In this section, we show some examples of the constructed QA pairs.

\begin{tcolorbox}[
  enhanced, breakable, title= \trqa — Sample 1,
  colback=grey, colframe=black!70, boxrule=0.8pt,
  colbacktitle=black!85, coltitle=white, fonttitle=\bfseries, title filled,
  sharp corners=all, arc=1.5mm, left=2mm, right=2mm, top=2mm, bottom=2mm
]

\begin{tcolorbox}[
  enhanced, breakable, title=Time Series Info ,
  colback=black!1, colframe=black!20, boxrule=0.5pt, title filled=false
]
\begin{minipage}[t]{1\linewidth}
  \includegraphics[width=\linewidth]{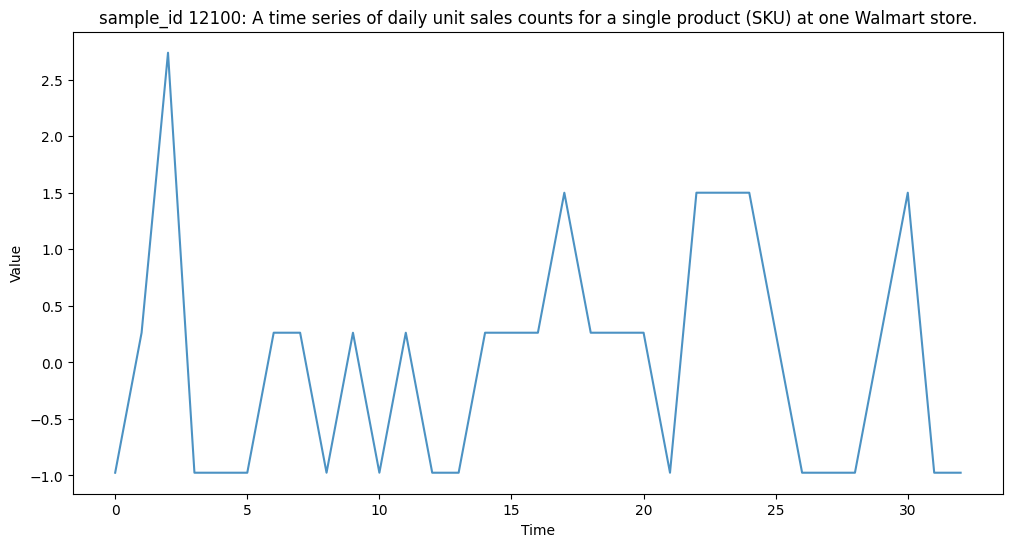}
\end{minipage}\hfill
\begin{minipage}[t]{\linewidth}\small
  \textbf{domain:} sales
\end{minipage}
\end{tcolorbox}

\vspace{0.6em}

\begin{tcolorbox}[
  enhanced, breakable, title=Question \& Answer,
  colback=black!1, colframe=black!20, boxrule=0.5pt, title filled=false
]
\small
\medskip
\textbf{question:} Which statement best describes the overall characteristics of this time series with regard to its
motif, variability, and trend? Respond \emph{ONLY} with the letter of the correct choice (A, B, C, or D).\\
Choices:
\begin{enumerate}[label=\Alph*:, leftmargin=*, itemsep=2pt]
  \item The time series has a simple motif, low variability, and lacks a consistent upward or downward trend.
  \item The time series features a complex motif, high variability, and an upward trend through the series.
  \item The time series has a constant value, no motif, and displays a strong downward trend.
  \item The time series contains random patterns with high variability and a significant upward trend.
\end{enumerate}

\textbf{question\_type:} multiple\_choices\quad
\textbf{task:} characterization

\medskip
\textbf{answer:} \textcolor{blue}{\textbf{A}}
\end{tcolorbox}

\end{tcolorbox}

\begin{tcolorbox}[
  enhanced, breakable, title=\trqa — Characterization Sample 1, colback=grey, colframe=black!70, boxrule=0.8pt,
  colbacktitle=black!85, coltitle=white, fonttitle=\bfseries, title filled,
  sharp corners=all, arc=1.5mm, left=2mm, right=2mm, top=2mm, bottom=2mm
]
\begin{tcolorbox}[
  enhanced, breakable, title=Time Series Info ,
  colback=black!1, colframe=black!20, boxrule=0.5pt, title filled=false
]
\begin{minipage}[t]{1\linewidth}
  \includegraphics[width=\linewidth]{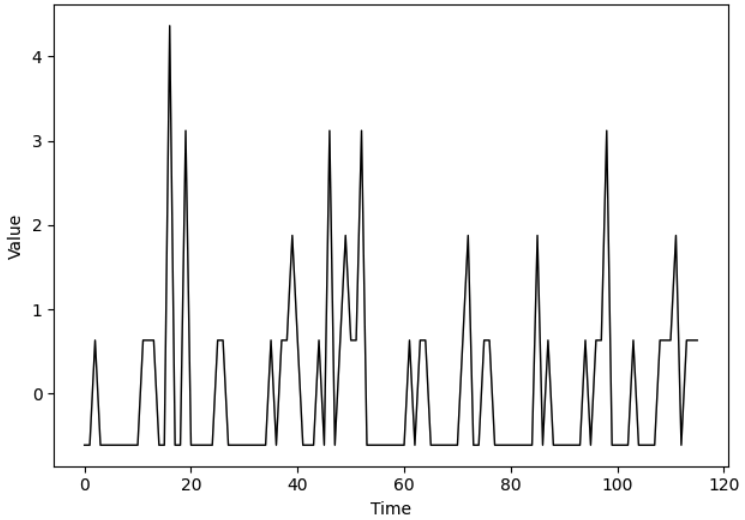}
\end{minipage}\hfill
\begin{minipage}[t]{\linewidth}\small
  \textbf{Description:} A numerical sequence of hourly aggregated Uber pickup counts (integer values) for a single New York City taxi zone, where each element represents the total number of pickups recorded in that zone during one hour.
  
  \textbf{Question Type:} TF
  \textbf{Domain:} transport
  \textbf{Dataset:} uber\_tlc\_hourly
\end{minipage}
\end{tcolorbox}
\vspace{0.6em}
\begin{tcolorbox}[
  enhanced, breakable, title=Question \& Answer,
  colback=black!1, colframe=black!20, boxrule=0.5pt, title filled=false
]
\small
\medskip
\textbf{Question:} Does the time series exhibit constant variance throughout, indicating no change in volatility, along with a clearly defined global shape?
Respond ONLY with the letter of the correct choice (T or F).\\
Choices:
\begin{enumerate}[leftmargin=*, itemsep=2pt]
  \item[T:] True.
  \item[F:] False.
\end{enumerate}
\medskip
\textbf{Answer:} \textcolor{blue}{\textbf{F}}
\end{tcolorbox}
\end{tcolorbox}
\vspace{5cm}
\begin{tcolorbox}[
  enhanced, breakable, title= \trqa — Characterization Sample 2, colback=grey, colframe=black!70, boxrule=0.8pt,
  colbacktitle=black!85, coltitle=white, fonttitle=\bfseries, title filled,
  sharp corners=all, arc=1.5mm, left=2mm, right=2mm, top=2mm, bottom=2mm
]
\begin{tcolorbox}[
  enhanced, breakable, title=Time Series Info ,
  colback=black!1, colframe=black!20, boxrule=0.5pt, title filled=false
]
\begin{minipage}[t]{1\linewidth}
  \includegraphics[width=\linewidth]{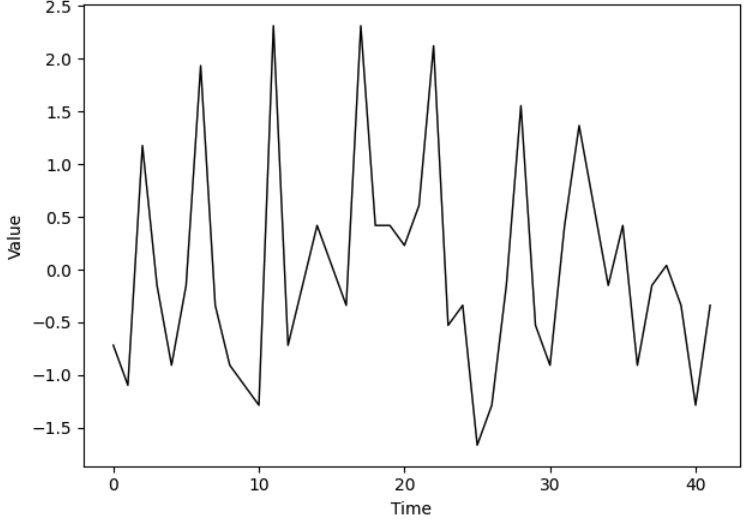}
\end{minipage}\hfill
\begin{minipage}[t]{\linewidth}\small
  \textbf{Description:} A time series of daily page‐view counts for a single English Wikipedia article.
  
  \textbf{Question Type:} MC
  \textbf{Domain:} web
  \textbf{Dataset:} wiki\_daily\_100k
\end{minipage}
\end{tcolorbox}
\vspace{0.6em}
\begin{tcolorbox}[
  enhanced, breakable, title=Question \& Answer,
  colback=black!1, colframe=black!20, boxrule=0.5pt, title filled=false
]
\small
\medskip
\textbf{Question:} Which of the following best describes a prominent motif and cyclic feature observed in the time series?
Respond ONLY with the letter of the correct choice (A, B, C, or D).\\
Choices:
\begin{enumerate}[label=\Alph*:, leftmargin=*, itemsep=2pt]
  \item A recurring increase in values every few days with a noticeable peak at day 11.
  \item A consistent downward trend over the entire period without any peaks.
  \item An irregular pattern with no identical sequences or cycles.
  \item A repeating cycle of gradual increase and sudden drop every ten days.
\end{enumerate}
\medskip
\textbf{Answer:} \textcolor{blue}{\textbf{A}}
\end{tcolorbox}
\end{tcolorbox}

\begin{tcolorbox}[
  enhanced, breakable, title= \trqa — Comparison Sample 1, colback=grey, colframe=black!70, boxrule=0.8pt,
  colbacktitle=black!85, coltitle=white, fonttitle=\bfseries, title filled,
  sharp corners=all, arc=1.5mm, left=2mm, right=2mm, top=2mm, bottom=2mm
]
\begin{tcolorbox}[
  enhanced, breakable, title=Time Series Info ,
  colback=black!1, colframe=black!20, boxrule=0.5pt, title filled=false
]
\begin{minipage}[t]{1\linewidth}
  \includegraphics[height=6cm, width=\linewidth]{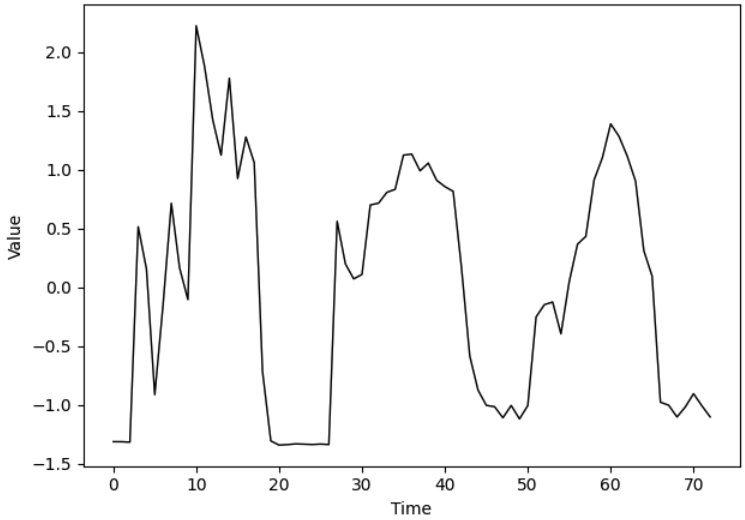}
  \includegraphics[height=6cm, width=\linewidth]{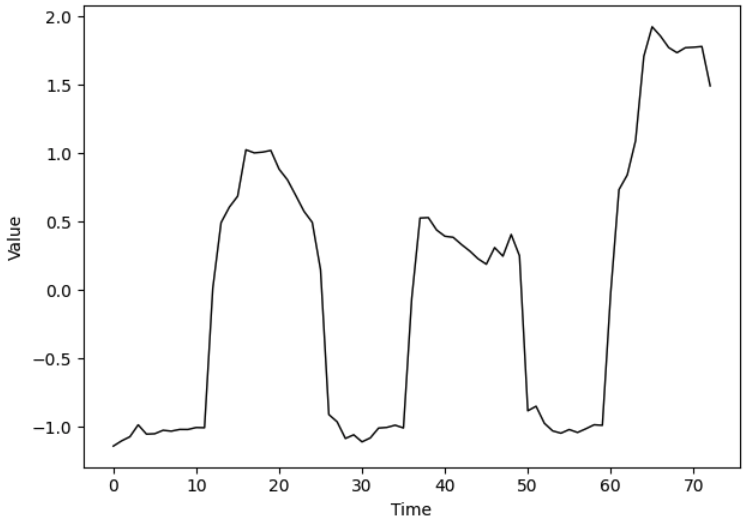}
\end{minipage}\hfill
\begin{minipage}[t]{\linewidth}\small
  \textbf{Description 1:} A time series of hourly electricity consumption measurements for the Rat building, representing one building\u2019s power usage. 
  
  \textbf{Description 2:} A time series of hourly electricity consumption measurements for the Rat building, representing one building\u2019s power usage.
  
  \textbf{Question Type:} TF
  \textbf{Domain:} energy
  \textbf{Dataset:} bdg\-2\_rat
\end{minipage}
\end{tcolorbox}
\vspace{0.6em}
\begin{tcolorbox}[
  enhanced, breakable, title=Question \& Answer,
  colback=black!1, colframe=black!20, boxrule=0.5pt, title filled=false
]
\small
\medskip
\textbf{Question:} Does time series 1 display any global upward trend over the entire period?\\
Choices:
\begin{enumerate}[leftmargin=*, itemsep=2pt]
  \item[T:] True.
  \item[F:] False.
\end{enumerate}
\medskip
\textbf{Answer:} \textcolor{blue}{\textbf{F}}
\end{tcolorbox}
\end{tcolorbox}

\begin{tcolorbox}[
  enhanced, breakable, title= \trqa — Comparison Sample 2, colback=grey, colframe=black!70, boxrule=0.8pt,
  colbacktitle=black!85, coltitle=white, fonttitle=\bfseries, title filled,
  sharp corners=all, arc=1.5mm, left=2mm, right=2mm, top=2mm, bottom=2mm
]
\begin{tcolorbox}[
  enhanced, breakable, title=Time Series Info ,
  colback=black!1, colframe=black!20, boxrule=0.5pt, title filled=false
]
\begin{minipage}[t]{1\linewidth}
  \includegraphics[height=6cm, width=\linewidth]{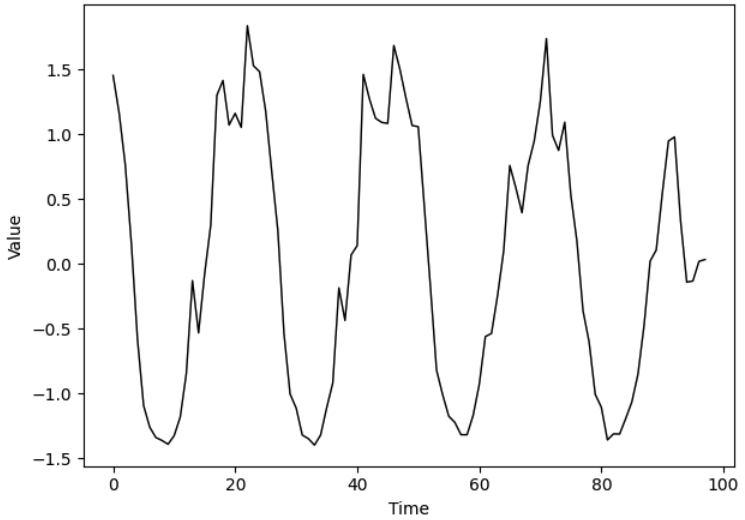}
  \includegraphics[height=6cm, width=\linewidth]{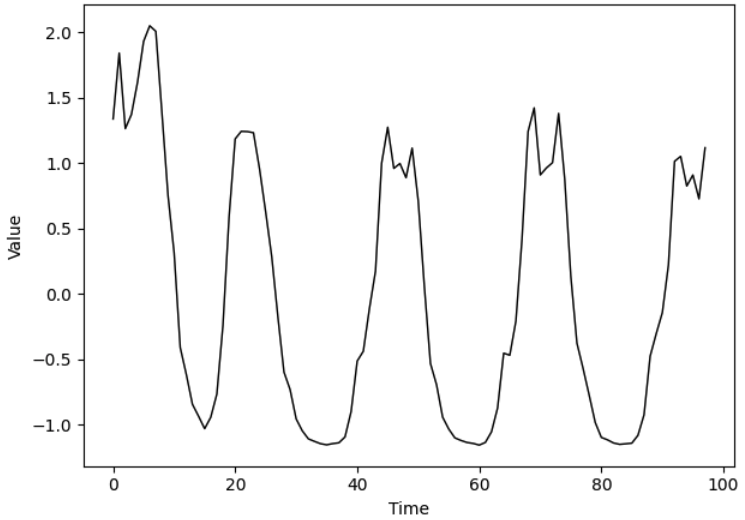}
\end{minipage}\hfill
\begin{minipage}[t]{\linewidth}\small
  \textbf{Description 1:} A time series of hourly pedestrian count measurements from a single sensor in Melbourne.

  \textbf{Description 2:} A time series of hourly pedestrian count measurements from a single sensor in Melbourne.
  
  \textbf{Question Type:} MC
  \textbf{Domain:} transport
  \textbf{Dataset:} pedestrian\_counts
\end{minipage}
\end{tcolorbox}
\vspace{0.6em}
\begin{tcolorbox}[
  enhanced, breakable, title=Question \& Answer,
  colback=black!1, colframe=black!20, boxrule=0.5pt, title filled=false
]
\small
\medskip
\textbf{Question:} Which time series displays stronger global stationarity, evident from its overall pattern smoothness without clear seasonal strength?
Respond ONLY with the letter of the correct choice (A, B, C, or D).\\
Choices:
\begin{enumerate}[label=\Alph*:, leftmargin=*, itemsep=2pt]
  \item Time series 1
  \item Time series 2
  \item Both have similar global stationarity
  \item Neither has strong global stationarity
\end{enumerate}
\medskip
\textbf{Answer:} \textcolor{blue}{\textbf{B}}
\end{tcolorbox}
\end{tcolorbox}

\begin{tcolorbox}[
  enhanced, breakable, title= \trqa — Data Transformation Sample 1, colback=grey, colframe=black!70, boxrule=0.8pt,
  colbacktitle=black!85, coltitle=white, fonttitle=\bfseries, title filled,
  sharp corners=all, arc=1.5mm, left=2mm, right=2mm, top=2mm, bottom=2mm
]
\begin{tcolorbox}[
  enhanced, breakable, title=Time Series Info ,
  colback=black!1, colframe=black!20, boxrule=0.5pt, title filled=false
]
\begin{minipage}[t]{1\linewidth}
  \includegraphics[height=6cm, width=\linewidth]{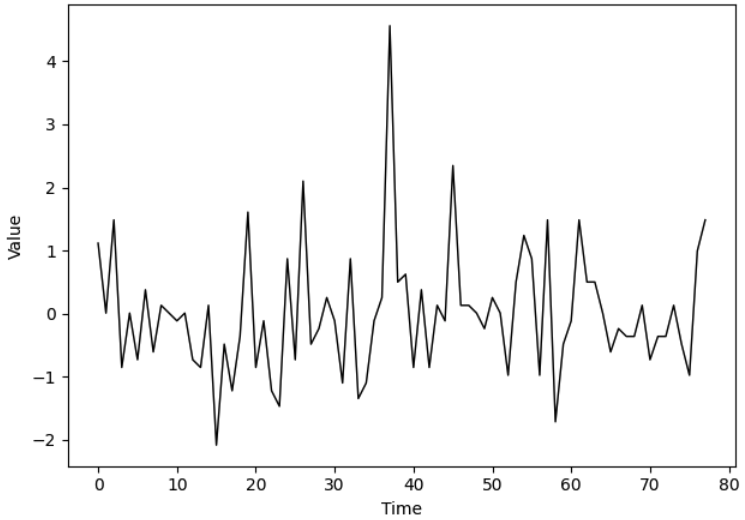}
\end{minipage}\hfill
\begin{minipage}[t]{\linewidth}\small
  \textbf{Description:} A time series of daily page\u2010view counts for a single English Wikipedia article.
  
  \textbf{Question Type:} MC
  \textbf{Domain:} web, energy, finance, finance
  \textbf{Dataset:} wiki\_daily\_100k, gfc12\_load, exchange\_rat, fred\_md
\end{minipage}
\end{tcolorbox}
\vspace{0.6em}
\begin{tcolorbox}[
  enhanced, breakable, title=Question \& Answer,
  colback=black!1, colframe=black!20, boxrule=0.5pt, title filled=false
]
\small
\medskip
\textbf{Question:} Which of the following choices is most likely the First Order Difference of the given time series?
Respond ONLY with the letter of the correct choice (A, B, C, or D).\\
Choices:
\begin{enumerate}[label=\Alph*:, leftmargin=*, itemsep=2pt]
  \item \parbox[t]{\linewidth}{%
    \vspace{0pt}%
    \includegraphics[width=\linewidth]{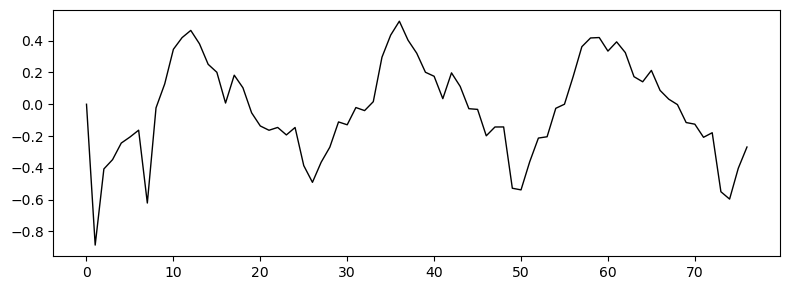}%
  }
 
  \item \parbox[t]{\linewidth}{%
    \vspace{0pt}%
    \includegraphics[width=\linewidth]{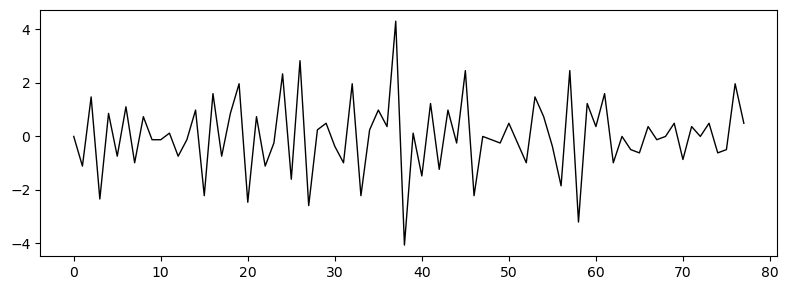}%
  }
 
  \item \parbox[t]{\linewidth}{%
    \vspace{0pt}%
    \includegraphics[width=\linewidth]{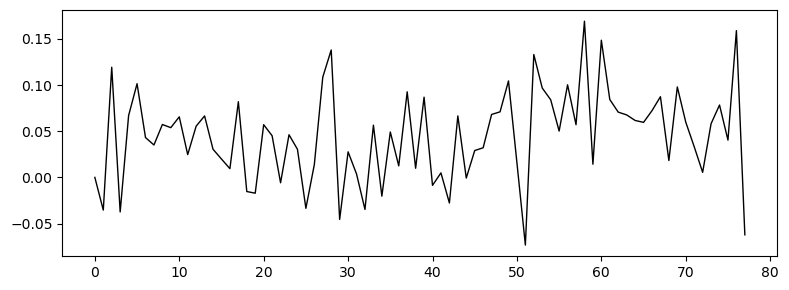}%
  }
  
  \item \parbox[t]{\linewidth}{%
    \vspace{0pt}%
    \includegraphics[width=\linewidth]{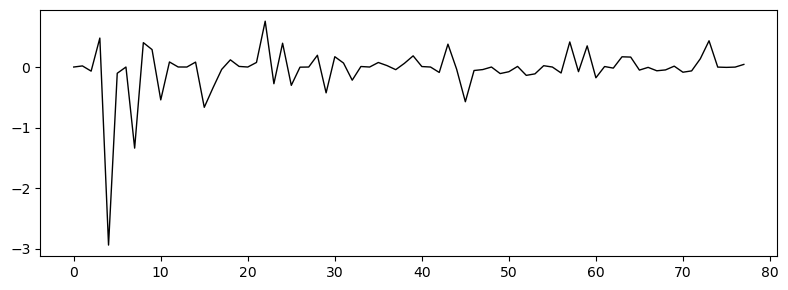}%
  }

\end{enumerate}
\medskip
\textbf{Answer:} \textcolor{blue}{\textbf{B}}
\end{tcolorbox}
\end{tcolorbox}
\begin{tcolorbox}[
  enhanced, breakable, title= \trqa — Data Transformation Sample 2, colback=grey, colframe=black!70, boxrule=0.8pt,
  colbacktitle=black!85, coltitle=white, fonttitle=\bfseries, title filled,
  sharp corners=all, arc=1.5mm, left=2mm, right=2mm, top=2mm, bottom=2mm
]
\begin{tcolorbox}[
  enhanced, breakable, title=Time Series Info ,
  colback=black!1, colframe=black!20, boxrule=0.5pt, title filled=false
]
\begin{minipage}[t]{1\linewidth}
  \includegraphics[height=6cm, width=\linewidth]{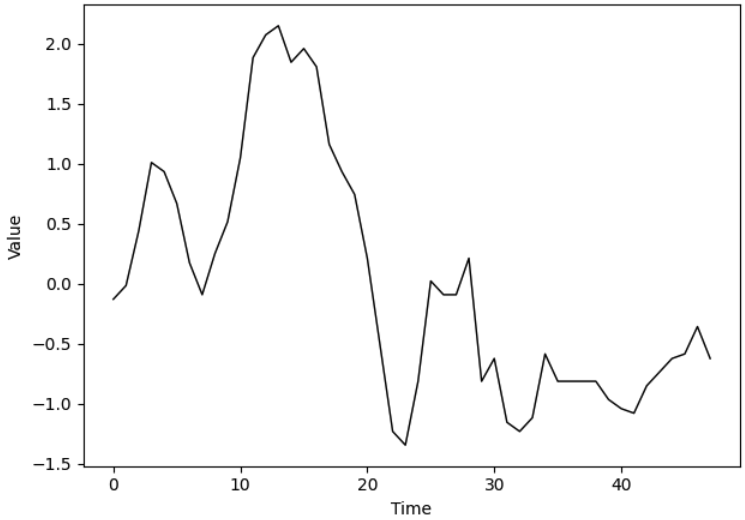}
\end{minipage}\hfill
\begin{minipage}[t]{\linewidth}\small
  \textbf{Description:} A time series of daily relative sunspot number measurements, where each value represents the quantified count of sunspot activity on the Sun\u2019s visible disk.
  
  \textbf{Question Type:} TF
  \textbf{Domain:} nature, energy
  \textbf{Dataset:} Nature\_sunspot, gfc12\_load
\end{minipage}
\end{tcolorbox}
\vspace{0.6em}
\begin{tcolorbox}[
  enhanced, breakable, title=Question \& Answer,
  colback=black!1, colframe=black!20, boxrule=0.5pt, title filled=false
]
\small
\medskip
\textbf{Question:} 
\\ Is the following sequence the First Order Difference of the given time series?
\begin{minipage}[t]{1\linewidth}
  \includegraphics[height=6cm, width=\linewidth]{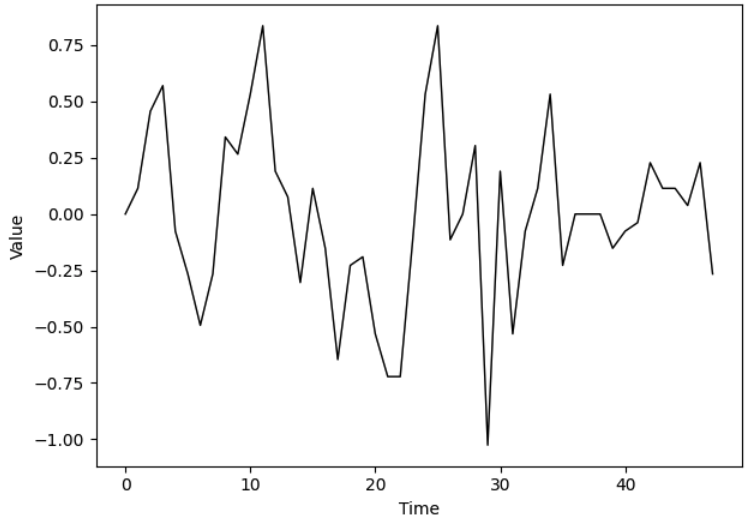}
\end{minipage}\hfill
Respond ONLY with the letter of the correct choice (T or F).\\
Choices:
\begin{enumerate}[leftmargin=*, itemsep=2pt]
  \item[T:] True.
  \item[F:] False.
\end{enumerate}
\medskip
\textbf{Answer:} \textcolor{blue}{\textbf{T}}
\end{tcolorbox}
\end{tcolorbox}


\begin{tcolorbox}[
  enhanced, breakable, title= \trqa — Temporal Relationship Sample 1, colback=grey, colframe=black!70, boxrule=0.8pt,
  colbacktitle=black!85, coltitle=white, fonttitle=\bfseries, title filled,
  sharp corners=all, arc=1.5mm, left=2mm, right=2mm, top=2mm, bottom=2mm
]
\begin{tcolorbox}[
  enhanced, breakable, title=Time Series Info ,
  colback=black!1, colframe=black!20, boxrule=0.5pt, title filled=false
]
\begin{minipage}[t]{1\linewidth}
  \includegraphics[width=\linewidth]{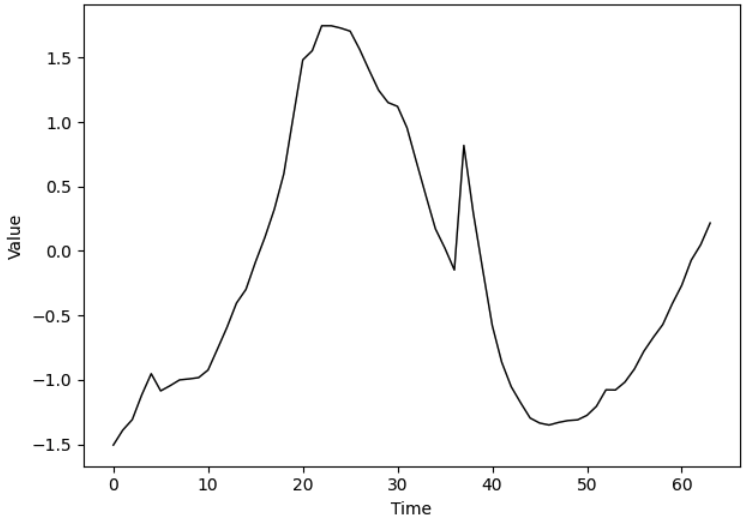}
\end{minipage}\hfill
\begin{minipage}[t]{\linewidth}\small

  \textbf{Question Type:} MC

\end{minipage}
\end{tcolorbox}
\vspace{0.6em}
\begin{tcolorbox}[
  enhanced, breakable, title=Question \& Answer,
  colback=black!1, colframe=black!20, boxrule=0.5pt, title filled=false
]
\small
\medskip
\textbf{Question:} Which of the following choices is most likely the future continuation of the given time series?
Respond ONLY with the letter of the correct choice (A, B, C, or D).\\
Choices:
\begin{enumerate}[label=\Alph*:, leftmargin=*, itemsep=2pt]
   \item \parbox[t]{\linewidth}{%
        \vspace{0pt}%
        \includegraphics[width=\linewidth]{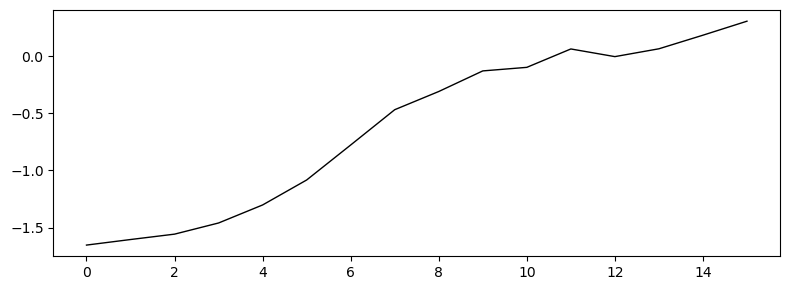}%
   }

  \item \parbox[t]{\linewidth}{%
        \vspace{0pt}%
        \includegraphics[width=\linewidth]{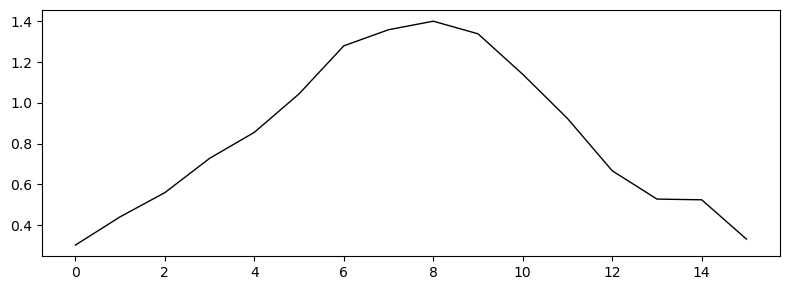}%
   }

  \item \parbox[t]{\linewidth}{%
        \vspace{0pt}%
        \includegraphics[width=\linewidth]{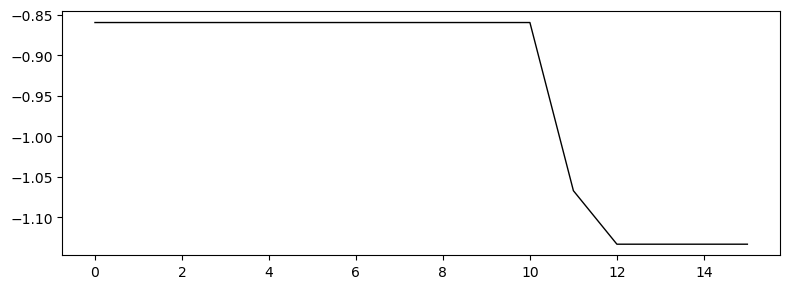}%
    }
  
  \item \parbox[t]{\linewidth}{%
        \vspace{0pt}%
        \includegraphics[width=\linewidth]{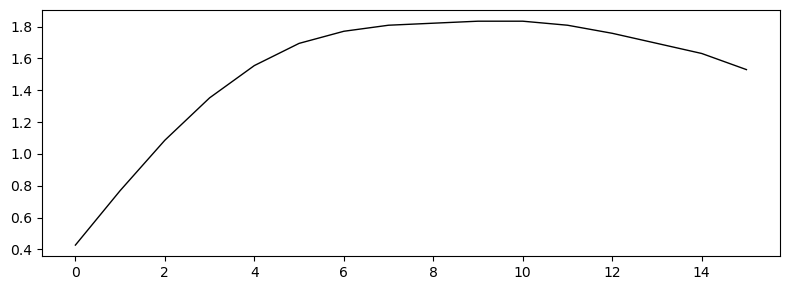}%
    }
\end{enumerate}
\medskip
\textbf{Answer:} \textcolor{blue}{\textbf{B}}
\end{tcolorbox}
\end{tcolorbox}
\begin{tcolorbox}[
  enhanced, breakable, title= \trqa — Temporal Relationship Sample 2, colback=grey, colframe=black!70, boxrule=0.8pt,
  colbacktitle=black!85, coltitle=white, fonttitle=\bfseries, title filled,
  sharp corners=all, arc=1.5mm, left=2mm, right=2mm, top=2mm, bottom=2mm
]
\begin{tcolorbox}[
  enhanced, breakable, title=Time Series Info ,
  colback=black!1, colframe=black!20, boxrule=0.5pt, title filled=false
]
\begin{minipage}[t]{1\linewidth}
  \includegraphics[width=\linewidth]{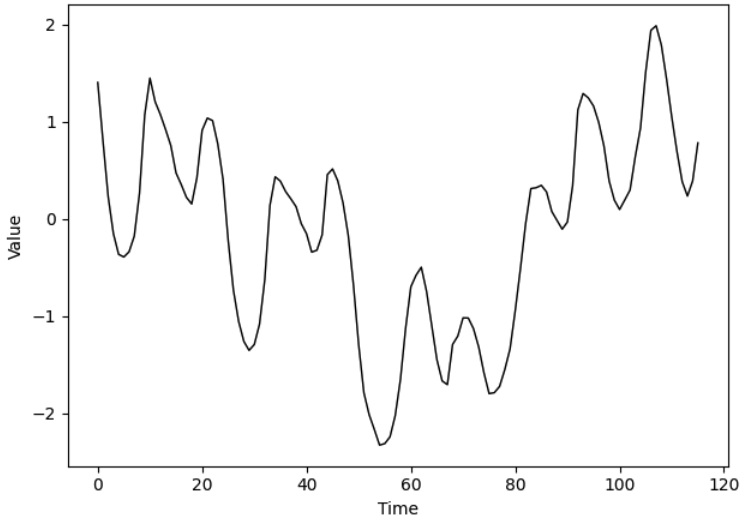}
\end{minipage}\hfill
\begin{minipage}[t]{\linewidth}\small
  
  \textbf{Question Type:} TF
\end{minipage}
\end{tcolorbox}
\vspace{0.6em}
\begin{tcolorbox}[
  enhanced, breakable, title=Question \& Answer,
  colback=black!1, colframe=black!20, boxrule=0.5pt, title filled=false
]
\small
\medskip
\textbf{Question:} \\Is the following patch the future continuation of the given time series?
\begin{minipage}[t]{1\linewidth}
  \includegraphics[width=\linewidth]{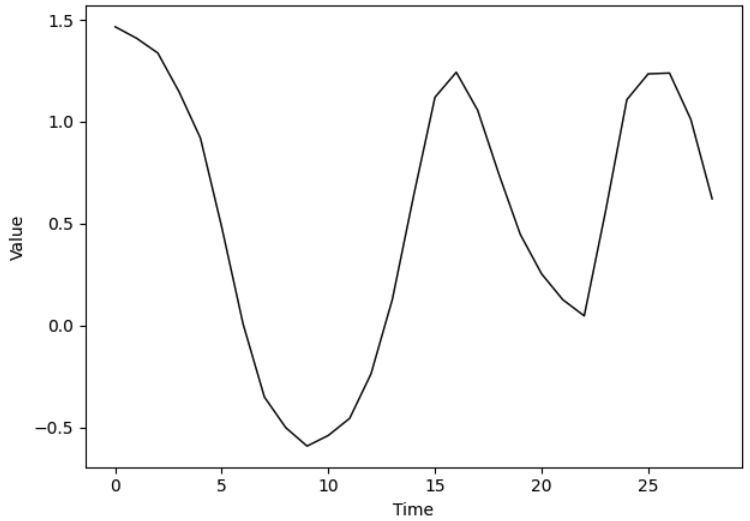}
\end{minipage}\hfill
Respond ONLY with the letter of the correct choice (T or F).\\
Choices:
\begin{enumerate}[leftmargin=*, itemsep=2pt]
  \item[T:] True.
  \item[F:] False.
\end{enumerate}
\medskip
\textbf{Answer:} \textcolor{blue}{\textbf{T}}
\end{tcolorbox}
\end{tcolorbox}

\vspace{5cm}
\begin{tcolorbox}[
  enhanced, breakable, title= \trqa — Temporal Relationship Sample 3, colback=grey, colframe=black!70, boxrule=0.8pt,
  colbacktitle=black!85, coltitle=white, fonttitle=\bfseries, title filled,
  sharp corners=all, arc=1.5mm, left=2mm, right=2mm, top=2mm, bottom=2mm
]
\begin{tcolorbox}[
  enhanced, breakable, title=Time Series Info ,
  colback=black!1, colframe=black!20, boxrule=0.5pt, title filled=false
]
\begin{minipage}[t]{1\linewidth}
  \includegraphics[width=\linewidth]{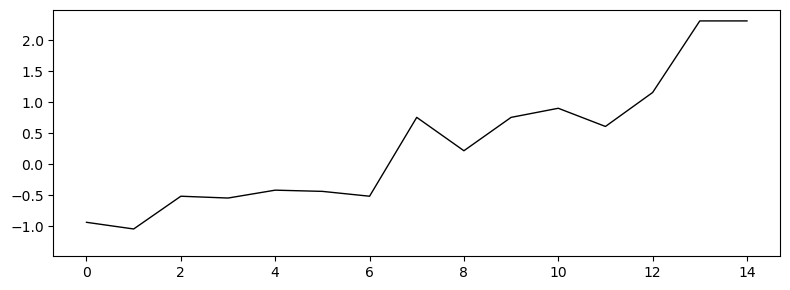}
\end{minipage}\hfill
\begin{minipage}[t]{\linewidth}\small
  
  \textbf{Question Type:} PZ
  \textbf{Domain:} finance
\end{minipage}
\end{tcolorbox}
\vspace{0.6em}
\begin{tcolorbox}[
  enhanced, breakable, title=Question \& Answer,
  colback=black!1, colframe=black!20, boxrule=0.5pt, title filled=false
]
\small
\medskip
\textbf{Question:} The given time series is the first patch of the sequence.
Below are the remaining patches, labeled as A, B, C, and D.
Arrange A, B, C, D in the correct order to reconstruct the original sequence.\\
Choices:
\begin{enumerate}[label=\Alph*:, leftmargin=*, itemsep=2pt]
  \item \parbox[t]{\linewidth}{%
    \vspace{0pt}%
    \includegraphics[width=\linewidth]{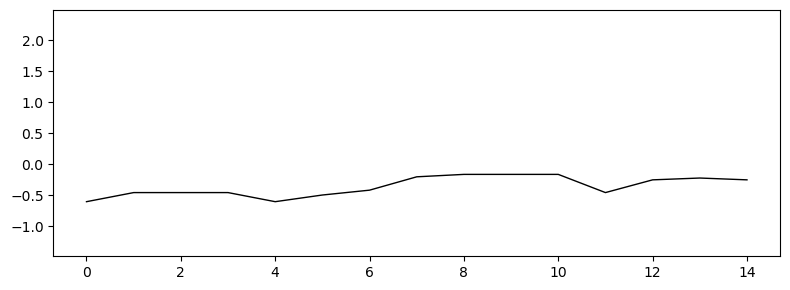}%
  }

  \item \parbox[t]{\linewidth}{%
    \vspace{0pt}%
    \includegraphics[width=\linewidth]{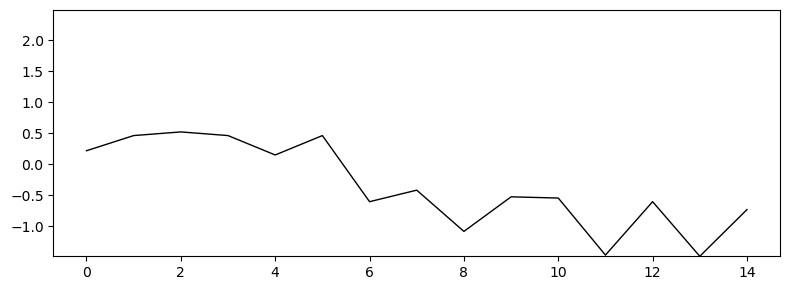}%
  }

  \item \parbox[t]{\linewidth}{%
    \vspace{0pt}%
    \includegraphics[width=\linewidth]{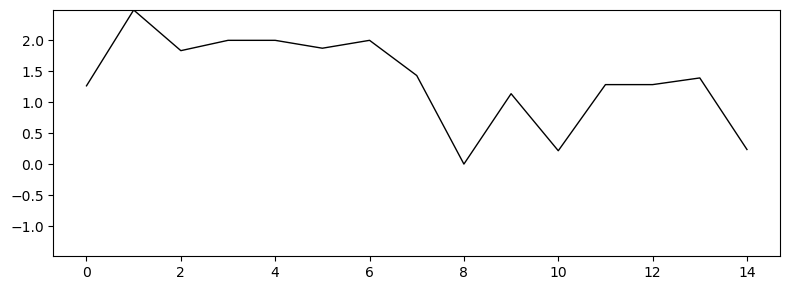}%
  }

  \item \parbox[t]{\linewidth}{%
    \vspace{0pt}%
    \includegraphics[width=\linewidth]{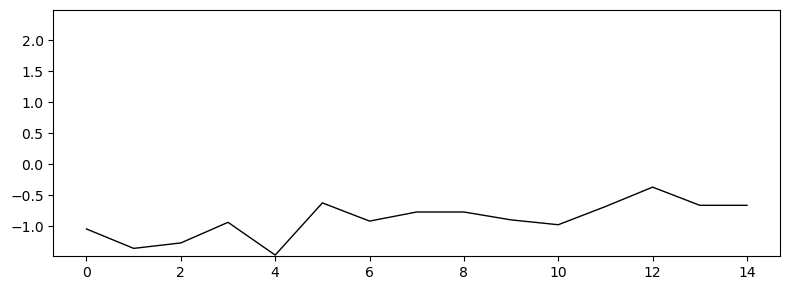}%
  }
  
\end{enumerate}
\medskip
\textbf{Answer:} \textcolor{blue}{\textbf{C,B,D,A}}
\end{tcolorbox}
\end{tcolorbox}

\begin{tcolorbox}[
  enhanced, breakable, title= \trqa — Anomaly Detection Sample 1, colback=grey, colframe=black!70, boxrule=0.8pt,
  colbacktitle=black!85, coltitle=white, fonttitle=\bfseries, title filled,
  sharp corners=all, arc=1.5mm, left=2mm, right=2mm, top=2mm, bottom=2mm
]
\begin{tcolorbox}[
  enhanced, breakable, title=Time Series Info ,
  colback=black!1, colframe=black!20, boxrule=0.5pt, title filled=false
]
\begin{minipage}[t]{1\linewidth}
  \includegraphics[width=\linewidth]{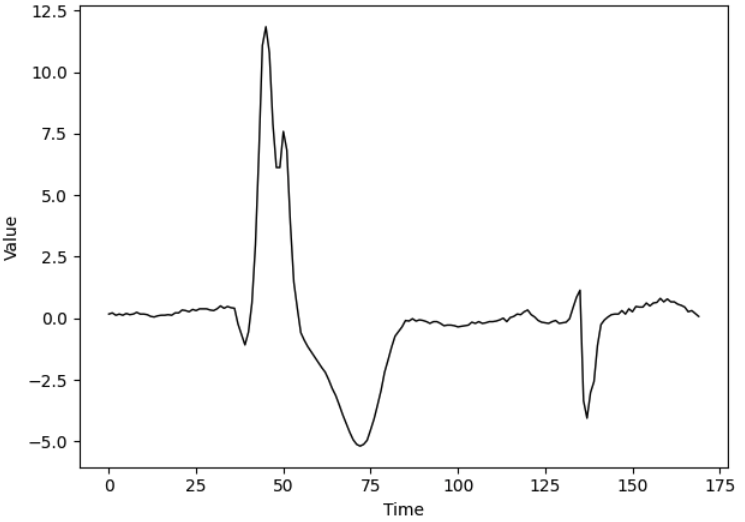}
\end{minipage}\hfill
\begin{minipage}[t]{\linewidth}\small
  \textbf{Description:} This is an electrocardiogram (ECG) time series, and the anomalies represent ventricular premature contractions. The ECG recordings were made using Del Mar Avionics model 445 two-channel reel-to-reel Holter recorders, and the analog signals were recreated for digitization using a Del Mar Avionics model 660 playback unit. The digitization rate (360 samples per second per channel) was chosen to accommodate the use of simple digital notch filters to remove 60 Hz (mains frequency) interference.
  
  \textbf{Question Type:} TF
  \textbf{Domain:} Healthcare
  \textbf{Dataset:} ECG
\end{minipage}
\end{tcolorbox}
\vspace{0.6em}
\begin{tcolorbox}[
  enhanced, breakable, title=Question \& Answer,
  colback=black!1, colframe=black!20, boxrule=0.5pt, title filled=false
]
\small
\medskip
\textbf{Question:} Determine whether the given time series contains anomalies.
Respond ONLY with the letter of the correct choice (T or F).\\
Choices:
\begin{enumerate}[leftmargin=*, itemsep=2pt]
  \item[T:] True.
  \item[F:] False.
\end{enumerate}
\medskip
\textbf{Answer:} \textcolor{blue}{\textbf{T}}
\end{tcolorbox}
\end{tcolorbox}

\begin{tcolorbox}[
  enhanced, breakable, title= \trqa — Classification Sample 1, colback=grey, colframe=black!70, boxrule=0.8pt,
  colbacktitle=black!85, coltitle=white, fonttitle=\bfseries, title filled,
  sharp corners=all, arc=1.5mm, left=2mm, right=2mm, top=2mm, bottom=2mm
]
\begin{tcolorbox}[
  enhanced, breakable, title=Time Series Info ,
  colback=black!1, colframe=black!20, boxrule=0.5pt, title filled=false
]
\begin{minipage}[t]{1\linewidth}
  \includegraphics[width=\linewidth]{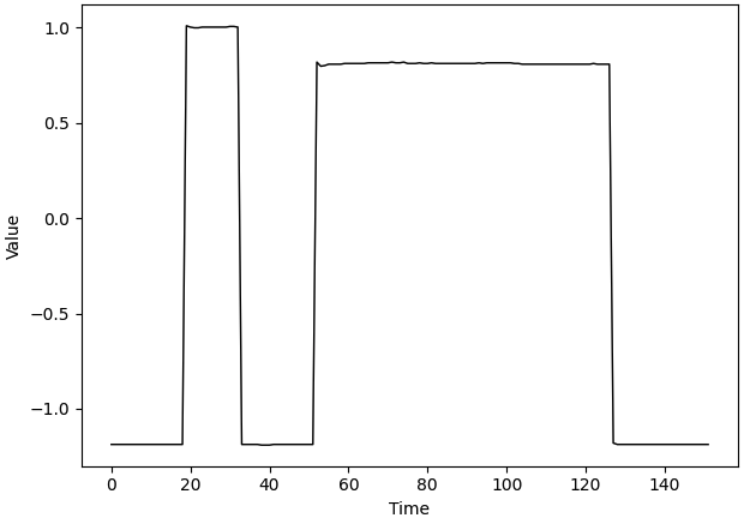}
\end{minipage}\hfill
\begin{minipage}[t]{\linewidth}\small
  \textbf{meta\_info:} This time series comes from a dataset capturing process control measurements recorded by individual sensors during the fabrication of silicon wafers in semiconductor manufacturing, providing data for monitoring and classifying normal and abnormal production processes.
  
  \textbf{Question Type:} MC
  \textbf{Domain:} manufacturing
  \textbf{Dataset:} UCR\_Classification\_Wafer
\end{minipage}
\end{tcolorbox}
\vspace{0.6em}
\begin{tcolorbox}[
  enhanced, breakable, title=Question \& Answer,
  colback=black!1, colframe=black!20, boxrule=0.5pt, title filled=false
]
\small
\medskip
\textbf{Question:} Classify the given time series into one of the categories below.
Respond ONLY with the letter of the correct choice (A, B).\\
Choices:
\begin{enumerate}[label=\Alph*:, leftmargin=*, itemsep=2pt]
  \item normal process
  \item abnormal process
\end{enumerate}
\medskip
\textbf{Answer:} \textcolor{blue}{\textbf{B}}
\end{tcolorbox}
\end{tcolorbox}

\newpage
\begin{figure*}[t]
    \centering
    \includegraphics[width=1\linewidth]{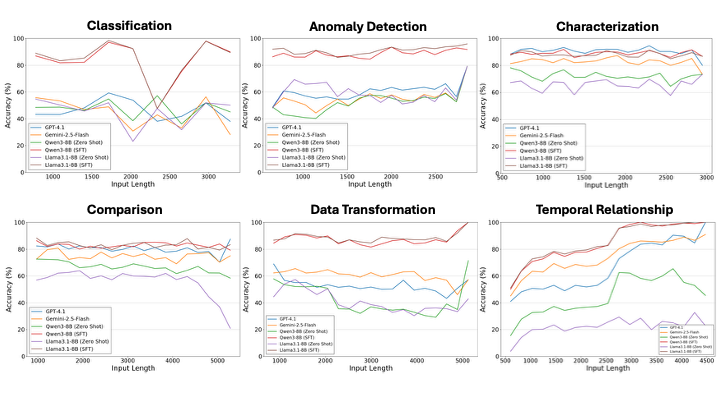}
    \caption{Input lengths vs. Accuracy by Tasks among six models.}
    \label{fig:input_length_accuracy}
\end{figure*}

\begin{figure*}[t]
    \centering
    \begin{subfigure}[t]{0.64\textwidth}
        \centering
        \includegraphics[width=\linewidth]{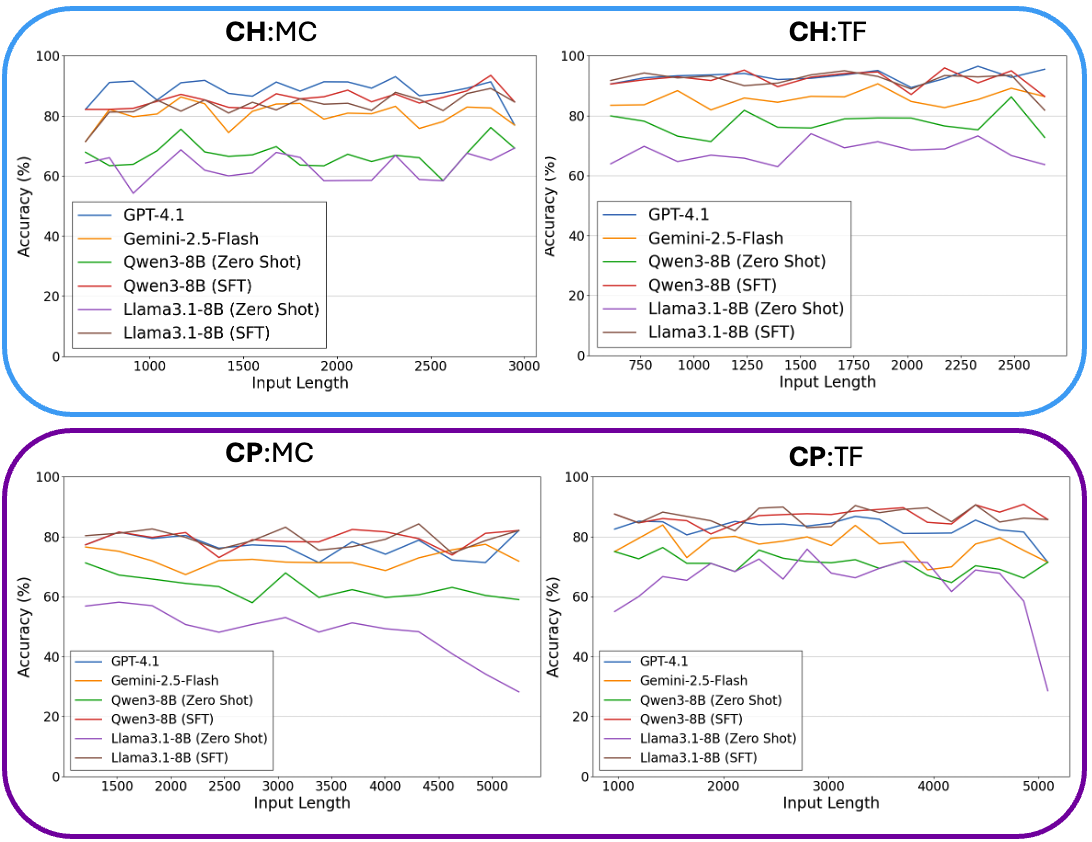}
        \label{fig:ch_cp_input_length_accuracy}
    \end{subfigure}
    \begin{subfigure}[t]{0.35\textwidth}
        \centering
        \includegraphics[width=\linewidth]{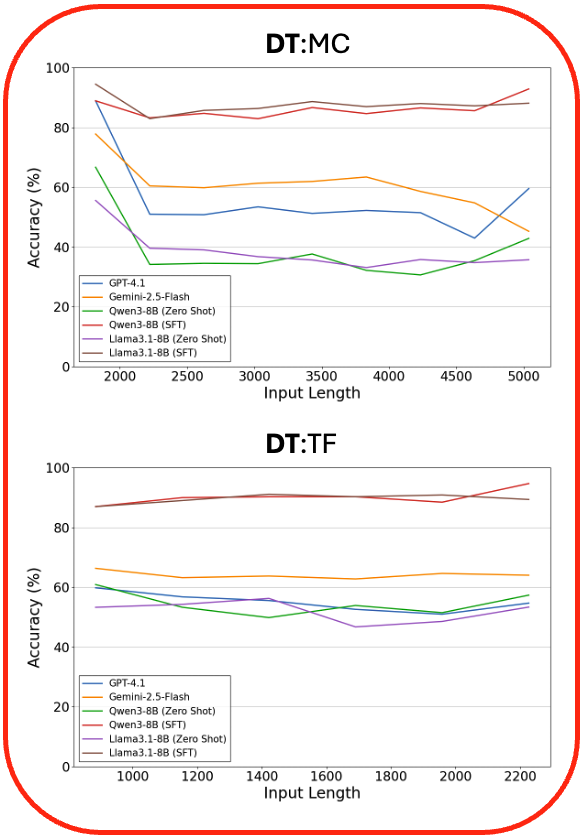}
        \label{fig:dt_input_length_accuracy}
    \end{subfigure}
    \begin{subfigure}[t]{0.99\textwidth}
        \centering
    \includegraphics[width=\linewidth]{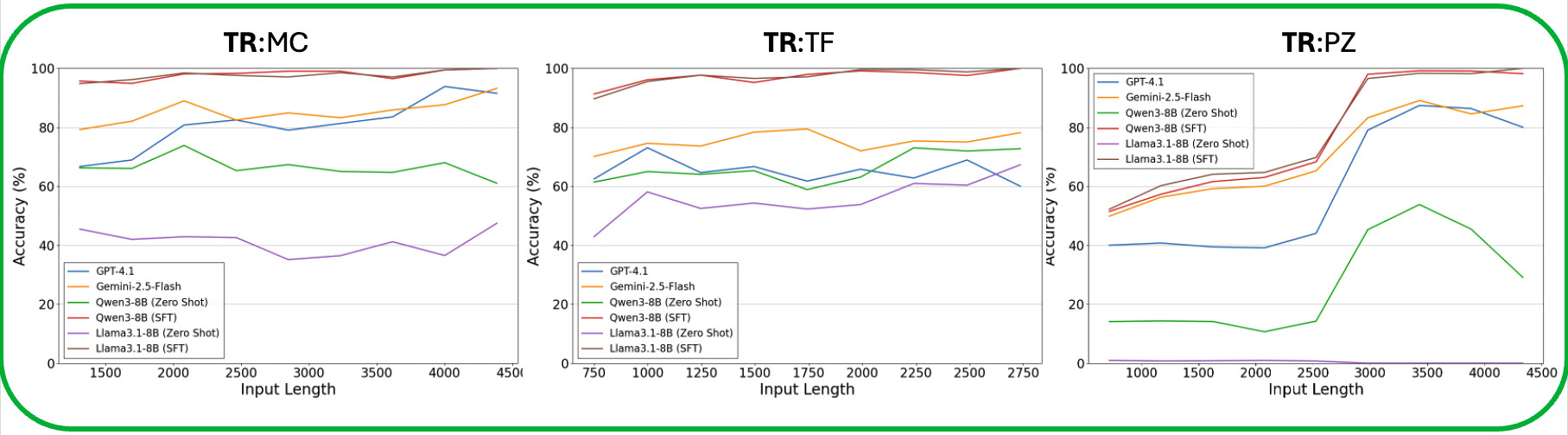}
        \label{fig:tr_input_length_accuracy}
    \end{subfigure}
    \caption{Input length vs. Accuracy by Question Types. CH, CP, DT, and TR denote Characterization, Comparison, Data Transformation, and Temporal Relationship. MC, TF, and PZ denote true-or-false, multiple-choice, and puzzling.}
    \label{fig:qt_input_length_accuracy}
\end{figure*}

\begin{table*}[t]
\centering
\scriptsize
\setlength{\tabcolsep}{3.5pt}
    \begin{tabular}{l|l|c|c|c|c}
        \hline
        \textbf{Group} & \textbf{Model} & \textbf{Topics} & \textbf{SubTopics} &
        \textbf{Characterization} &
        \textbf{Comparison} \\
        \hline
        \multirow{12}{*}{\textbf{Zero Shot}}
         & \multirow{3}{*}{GPT-4.1}         & 1 & 1 & 89.53 & 80.39 \\ 
         &                                  & 2 & 2 & 91.75 & 80.87 \\ 
         &                                  & 3 & 3 & 92.10 & 79.51 \\ 
         \cline{2-6}
         & \multirow{3}{*}{Gemini-2.5-Flash} & 1 & 1 & 83.68 & 77.57 \\ 
         &                                  & 2 & 2 & 83.01 & 74.57 \\ 
         &                                  & 3 & 3 & 83.11 & 72.62 \\ 
         \cline{2-6} 
         & \multirow{3}{*}{Qwen3-8B}         & 1 & 1 & 69.14 & 67.48 \\ 
         &                                  & 2 & 2 & 72.35 & 66.38 \\ 
         &                                  & 3 & 3 & 74.86 & 67.08 \\ 
         \cline{2-6}
         & \multirow{3}{*}{LLaMA3.1-8B}      & 1 & 1 & 63.85 & 58.93 \\ 
         &                                  & 2 & 2 & 66.16 & 58.75 \\ 
         &                                  & 3 & 3 & 65.53 & 58.71 \\ 
        \hline
        \multirow{6}{*}{\textbf{Instruction Tuning}}
         & \multirow{3}{*}{Qwen3-8B (SFT)}    & 1 & 1 & 86.22 & 82.48 \\ 
         &                                  & 2 & 2 & 89.74 & 82.50 \\ 
         &                                  & 3 & 3 & 90.57 & 83.15 \\ 
         \cline{2-6}
         & \multirow{3}{*}{LLaMA3.1-8B (SFT)} & 1 & 1 & 86.02 & 84.81 \\ 
         &                                  & 2 & 2 & 88.89 & 81.84 \\ 
         &                                  & 3 & 3 & 89.14 & 82.18 \\ 
        \hline
    \end{tabular}
\caption{Number of topics and subtopics v.s. Score}
\label{tab:subtopic-topic-analysis}
\end{table*}
\section{Experiment Analysis}
We conducted an in-depth analysis of results from the selected Large Language Models. Specifically, our analysis is divided into three major categories: \textbf{Accuracy Correlate Analysis}, \textbf{Task-Specific Analysis}, and \textbf{Case Study}. For each analysis, we selected models from both commercial and open-source families. In particular, we chose the two best-performing models from Table~\ref{tab:main_results} evaluated on our \trqa Benchmark—namely, GPT-4.1, Gemini 2.5 Flash, LLaMA3-8B, and Qwen3-8B. For LLaMA3.1-8B and Qwen3-8B, we analyzed both the zero-shot and instruction tuned models, resulting in a total of six models considered in our analysis.

\subsection{Accuracy Correlate Analysis}
\label{appendix:accuracy_correlate_analysis}
In this category, we primarily examined how model accuracy or overall score correlates with various factors, including input length, the topics and subtopics used in time-series question generation, and the influence of domain differences on model performance.

\noindent \textbf{Input Length v.s. Accuracy}. 
To understand how input length impacts model accuracy, we conducted a detailed analysis comparing the length of each input with its corresponding accuracy. Specifically, the input length is calculated as \textit{len(ts + description + domain + dataset + task + question\_type + question)} with \textit{String} type. The results are visualized in Figure~\ref{fig:input_length_accuracy}. Each plot may contain input length starting and ending at different length as each task contains questions with different lengths. Across all six models and five tasks, excluding the Temporal Relation task, we observe a consistent trend that longer questions with greater input length generally result in lower accuracy and weaker overall model performance. However, the Temporal Relation task exhibits the opposite behavior, where accuracy improves with increasing input length. 

To understand this discrepancy, we conducted a detailed analysis of the four advanced analysis tasks \textit{(Characterization, Comparison, Data Transformation, Temporal Relation)} in our proposed \trqa Benchmark, focusing on how different question types \textit{(MC, TF, PZ)} and their corresponding input lengths correlate with model accuracy. The results are visualized in Figure~\ref{fig:qt_input_length_accuracy}. The results indicate that for all four advanced analysis tasks, MC and TF question types show a decline in accuracy with increasing input length, whereas the newly proposed PZ type exhibits the opposite trend. 
This implies that the model is actively using global contexts, all time series segments, to deduce the correct chronological order for answering PZ type question, which confirms that the model is engaging in deductive reasoning rather than local pattern matching. This proves that PZ type question is a rigorous probe for \textit{Global Causal Reasoning}. Consequently, models whose accuracy improves with input length likely demonstrate a stronger ability to reason directly over time-series patterns.\\

\begin{figure*}[t]
    \centering
    \includegraphics[width=1\linewidth]{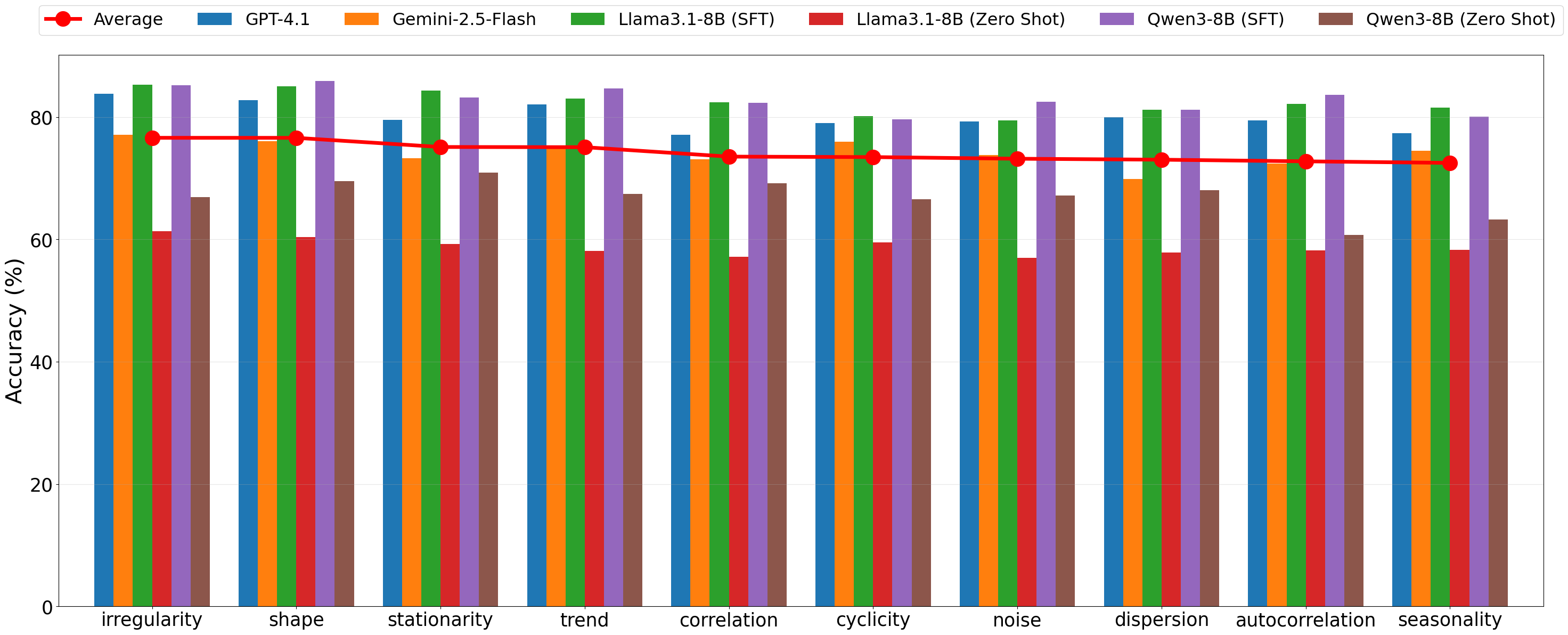}
    \caption{Topics vs. Accuracy of Comparison Task.}
    \label{fig:topics123}
\end{figure*}

\noindent \textbf{Topics \& Subtopics v.s. Accuracy.} In our proposed \trqa Benchmark, tasks such as Characterization and Comparison include questions generated by prompting GPT to select topics and subtopics from a predefined list~\ref{table:topics}. 

To understand how the complexity of topics and subtopics influences model performance, we analyzed the relationship between the number of topics and subtopics used in each question and the corresponding model accuracy. Specifically, we examined how varying topic and subtopic counts affect the model’s ability to reason across different levels of conceptual complexity. In our benchmark, each question contains between one to three topics, indicating that one to three distinct topics are considered during question generation, and between one to three associated subtopics depending on the selected topics. The results are summarized in Table~\ref{tab:subtopic-topic-analysis}. Based on the results, we observe that the complexity of questions with varying number of topics and subtopics don't have direct impact on model accuracy. The absence of significant performance differences across varying topics and subtopics indicates that the \trqa Benchmark is largely unbiased, suggesting that models do not rely on topic-level content from the question, but depend mainly on their time-series analytical capability. 

In addition, we further analyzed the difficulty of individual topics by examining how different topic combinations influence model performance across tasks. For questions containing more than one topic, each question was expanded into multiple rows, allowing us to isolate the accuracy associated with each topic across all questions that included it. Comparing the average accuracy of questions containing each topic allowed us to identify which topics posed greater challenges, especially when combined with others. The results of questions from Comparison Task are visualized in Figure~\ref{fig:topics123}. Generally, we found that questions with topics such as seasonality, autocorrelation, dispersion, and noise are harder for model.

\noindent \textbf{Domain v.s. Accuracy.} In the \trqa Benchmark, both questions and descriptions are generated from the associated dataset and domain.
\begin{table*}[t]
\centering
\scriptsize
\setlength{\tabcolsep}{3pt}
    \begin{tabular}{l|l|c|c|c|c|c|c|c|c|c|c|c|c|c}
        \hline
        \textbf{Group} & \textbf{Model} & \textbf{Mf} & \textbf{ES} &
        \textbf{Rbt} &
        \textbf{Bm} &
        \textbf{It} &
        \textbf{Eg} &
        \textbf{Hc} &
        \textbf{TP} & 
        \textbf{Nt} &
        \textbf{Fc} &
        \textbf{S}&
        \textbf{W}&
        \textbf{Sc}\\
        \hline
        \multirow{4}{*}{\textbf{Zero Shot}}
         & \multirow{1}{*}{GPT-4.1}         & 84.29 & 72.31 & \underline{38.91} & 49.93 & 48.56 & 62.91 & 64.18 & 66.07 & 64.82 & 67.49 & 68.73 & 67.21 & \textbf{26.95}\\ 
         \cline{2-15}
         & \multirow{1}{*}{Gemini-2.5-Flash} & 76.58 & 59.70 & 57.59 & 49.78 & \underline{48.28} & 69.44 & 70.40 & 71.55 & 69.52 & 73.88 & 70.12 & 66.02 & \textbf{14.55}\\ 
         \cline{2-15} 
         & \multirow{1}{*}{Qwen3-8B}         & 83.85 & 53.09 & 58.37 & 50.02 & \underline{48.56} & 50.45 & 50.91 & 49.78 & 50.60 & 51.08 & 50.99 & 50.47 & \textbf{15.49} \\ 
         \cline{2-15}
         & \multirow{1}{*}{LLaMA3.1-8B}      & 73.45 & 62.42 & 59.92 & 56.48 & 47.75 & 42.39 & 42.18 & 41.74 & 41.16 & 40.86 & 40.79 & \underline{40.07} & \textbf{21.88} \\ 
        \hline
        \multirow{2}{*}{\textbf{Instruction Tuning}}
         & \multirow{1}{*}{Qwen3-8B}    & 95.05 & 94.93 & 84.05 & 85.44 & 86.21 & 85.26 & 85.15 & 84.84 & 83.32 & 85.25 & \textbf{82.48} & \underline{82.49} & 98.31 \\ 
         \cline{2-15}
         & \multirow{1}{*}{LLaMA3.1-8B} & 96.22 & 96.23 & 87.94 & 87.34 & 92.00 & 86.63 & 85.80 & 85.4 & 84.28 & 85.83 & \textbf{82.23} & \underline{83.01} & 98.03\\ 
        \hline
    \end{tabular}
    \caption{Domain v.s. Accuracy. Mf denotes Manufacturing. ES denotes Environment Sensing. Rbt denotes Robotics. Bm denotes Biomedical. Eg denotes Energy. Hc denotes Healthcare. Tp denotes Transport. Nt denotes Nature. Fc denotes Finance. S denotes Sales. W denotes Web. Sc denotes Synthetic. The lowest and second-lowest results for each model are highlighted in \textbf{bold} and \underline{underlined}, respectively.}
    \label{tab:domain_accuracy}
\end{table*}

\begin{table*}[t]
    \centering
    \scriptsize
    \begin{tabular}{c | l | c c c | c c c}
        \toprule
        \multirow{2}{*}{\textbf{Group}} & \multirow{2}{*}{\textbf{Model}} & \multicolumn{3}{c|}{\textbf{MC}} & \multicolumn{3}{c}{\textbf{TF}} \\
        \cline{3-8} 
        & & FT & WT & FOD & FT & WT & FOD \\
        \midrule
        \multirow{4}{*}{\makecell{Zero\\Shot}}
        & GPT-4.1         & 26.32 & 35.39 & 91.90  & 51.36 & 51.64 & 59.81  \\
        & Gemini-2.5-Flash& 27.97 & 53.19 & 100.00 & 50.25 & 53.59 & 85.90 \\
        \cline{2-8} 
        & Qwen3-8B        & 9.06 & 28.40 & 66.4 & 52.57 & 52.05 & 52.66 \\
        & LLaMA3.1-8B     & 24.07 & 23.87 & 61.70 & 52.17 & 48.87 & 54.50 \\
        
        \midrule

        \multirow{2}{*}{\makecell{Instruction\\Tuning}}
        & Qwen3-8B            & 67.93 & 87.55 & 100.00 & 80.02 & 99.90 & 89.14 \\
        & LLaMA3.1-8B         & 71.83 & 88.79 & 99.70 & 82.54 & 89.24 & 98.36 \\
        \bottomrule
    \end{tabular}
    \caption{Analysis of Data Transformation Task. MC and TF denote multiple-choice and true-or-false, respectively. FT, WT, and FOD denote Fourier Transform, Wavelet Transform, and First-Order Differencing. We evaluate the accuracy on MC and TF questions from Data Transformation Task for each of the three transform operators.}
    \label{tab:analysis_datatransformation}
\end{table*} 

\begin{table*}[h]
\centering
\scriptsize
\setlength{\tabcolsep}{3.5pt}
    \begin{tabular}{l|l|c|c|c|c|c|c|c}
            \hline
            \textbf{Group} & \textbf{Model} & \textbf{Finance} & \textbf{Healthcare} &
            \textbf{Transport} &
            \textbf{Sales} &
            \textbf{Energy} &
            \textbf{Nature} &
            \textbf{Web} \\
            \hline
            \multirow{4}{*}{\textbf{Zero Shot}}
             & \multirow{1}{*}{GPT-4.1}         & 62.22 & 57.75 & 55.86 & 52.62 & 52.53 & \underline{48.87} & \textbf{46.53} \\ 
             \cline{2-9}
             & \multirow{1}{*}{Gemini-2.5-Flash} & 76.59 & 80.12 & 76.65 & \underline{66.54} & 76.95 & 72.46 & \textbf{63.51} \\ 
             \cline{2-9} 
             & \multirow{1}{*}{Qwen3-8B}         & 27.81 & 27.76 & 24.36 & 22.87 & 24.32 & \underline{21.43} & \textbf{17.90} \\ 
             \cline{2-9}
             & \multirow{1}{*}{LLaMA3.1-8B}      & \textbf{0.77} & \underline{0.78} & 0.98 & 0.94 & 0.88 & 1.25 & 0.92 \\ 
            \hline
            \multirow{2}{*}{\textbf{Instruction Tuning}}
             & \multirow{1}{*}{Qwen3-8B}    & 73.31 & 77.11 & 72.54 &  \underline{61.03} & 74.05 & 68.92 & \textbf{58.61} \\ 
             \cline{2-9}
             & \multirow{1}{*}{LLaMA3.1-8B} & 75.25 & 77.50 & 72.22 & \underline{61.80} & 75.80 & 71.16 & \textbf{60.86} \\ 
            \hline
        \end{tabular}
    \caption{Domain v.s. Accuracy of the PZ question type in the Temporal Relationship task. The lowest and second-lowest results for each model are highlighted in \textbf{bold} and \underline{underlined}, respectively.}
    \label{tab:pz_domain_accuracy}
\end{table*}

To evaluate the influence of such contextual information, we conducted an in-depth analysis of how domain variation impacts overall model accuracy on \trqa Benchmark. The results are summarized in Table~\ref{tab:domain_accuracy}. Our analysis reveals that questions from domains including Synthetic, IT, Robotics, and Web pose greater challenges to models under the zero-shot setting, while questions from Sales and Web domains remain the most difficult after instruction tuning. Notably, the questions from Synthetic domain, which initially produced the lowest accuracies across all models, show the most substantial improvement after instruction tuning, achieving the highest scores among all domains. However, Web-related questions persist as difficult cases, indicating that domain-specific complexities in this category are not fully mitigated by instruction tuning.

\subsection{Task Specific Analysis}
\label{appendix:task_specific_analysis}
In this category, we examined how each model performed across the tasks proposed in our \trqa Benchmark. Specifically, we focused on the 3 analysis tasks: Data Transformation, Temporal Relationship, and Comparison.\\

\noindent \textbf{Data Transformation}. 
We analyze model performance on the Data Transformation task, which is designed to evaluate a model's understanding of three transformation operators: Fourier Transform (FT), Wavelet Transform (WT), and First-Order Differencing (FOD). 
For each operator, we assess performance by measuring the accuracy on both MC and TF question formats. 
As shown in Table~\ref{tab:analysis_datatransformation}, for zero-shot evaluation, our key finding highlights a limitation in which both commercial and open-source models fail to provide accurate answers, except of FOD. 
In  contrast, our instruction-tuned models show a better performance, achieving high accuracy across all tasks. 
However, FT is still very challenging even after instruction tuning. 

To explain our findings, we attribute this systematic performance disparity to two primary factors: the scope of temporal dependency and arithmetic complexity. As shown in Table~\ref{tab:analysis_datatransformation}, there is a clear performance degradation trend ($FOD > WT > FT$). This performance degradation is likely due to 3 reasons. (1) FOD relies solely on adjacent time steps ($x_t - x_{t-1}$), aligning well with the local attention capabilities of Transformers. (2) WT requires reasoning over localized windows in both time and frequency. As the dependency scope widens beyond immediate neighbors, model performance drops. (3) FT necessitates aggregating information from the entire sequence to determine frequency components. This global arithmetic reasoning is inherently challenging for LLMs' next-token prediction paradigm, resulting in the lowest performance. The results systematically validate that current LLMs struggle with tasks requiring global aggregation and complex arithmetic compared to robust local pattern matching, which also explains the results shown in Table~\ref{tab:analysis_datatransformation}.\\

\noindent \textbf{Temporal Relationship}. 
We analyzed model performance on the Temporal Relationship task, focusing specifically on our newly proposed Puzzling (PZ) question type. Beyond the input length versus accuracy analysis previously presented in Figure~\ref{fig:qt_input_length_accuracy}, we further examined how domain-level information influences model performance on Puzzling questions and the nature of model errors. 

\textit{Domain-Level Analysis.} The results are summarized in Table~\ref{tab:pz_domain_accuracy}. The results show that the Web domain remains the most challenging for Puzzling questions across both zero-shot and instruction-tuning settings. Sales and Nature also exhibit lower accuracies, with Sales remaining difficult even after instruction-tuning. This indicates that domains such as Web and Sales impose greater temporal analysis difficulty on models, which is consistent with our findings from the overall domain vs. score analysis presented in Table~\ref{tab:domain_accuracy}. 


\begin{table*}[h]
    \centering
    \scriptsize
    \setlength{\tabcolsep}{3.5pt} 
    \begin{tabular}{l|l|l|ccccccc}
        \toprule
        \textbf{Group} & \textbf{Model} & \textbf{Metric} & \textbf{Finance} & \textbf{Healthcare} & \textbf{Transport} & \textbf{Sales} & \textbf{Energy} & \textbf{Nature} & \textbf{Web} \\
        \midrule
        \multirow{6}{*}{\textbf{Instruction Tuning}} 
        & \multirow{3}{*}{Qwen3-8B} 
          & $D_{gt}$   & 1.21 & 1.84 & 1.48 & 2.50 & 0.93 & 0.90 & 2.57 \\
        & & $D_{pred}$ & 1.67 & 1.81 & 1.56 & 2.00 & 1.28 & 1.40 & 2.31 \\
        & & Gap & -0.46 & 0.03 & -0.08 & \textbf{0.50} & -0.35 & -0.50 & \underline{0.26} \\
        \cmidrule{2-10}
        & \multirow{3}{*}{LLaMA3.1-8B} 
          & $D_{gt}$   & 1.30 & 1.97 & 1.51 & 2.52 & 0.94 & 0.91 & 2.59 \\
        & & $D_{pred}$ & 1.51 & 1.84 & 1.46 & 2.02 & 1.18 & 1.20 & 2.14 \\
        & & Gap & -0.21 & 0.13 & 0.05 & \textbf{0.50} & -0.24 & -0.29 & \underline{0.45} \\
        \bottomrule
    \end{tabular}
    \caption{Analysis of Smoothness Bias in PZ question type. For each domain, we report the boundary distance of Ground Truth ($D_{gt}$), Predicted ($D_{pred}$), and the Smoothness Gap ($Gap$). The largest and second-largest Gap for each model are highlighted in \textbf{bold} and \underline{underlined}, indicating the model is over-smoothing.}
    \label{tab:smoothness_bias_detailed}
\end{table*}
\textit{Error Analysis: Smoothness Bias.} To understand why models struggle in these specific domains (Web, Sales), we further analyzed the boundary consistency of incorrect predictions using the instruction-tuned LLaMA3.1-8B and Qwen3-8B. We calculated the "Smoothness Gap", defined as the difference between the boundary distance of the Ground Truth sequence \textit{$D_{gt}$} and the model Predicted sequence \textit{$D_{pred}$}. Here, the boundary distance is calculated as the Euclidean distance between the last time step of a preceding patch and the first time step of its succeeding patch. We found that in the challenging domains identified above, the models consistently constructed sequences where the boundary transitions were smoother than the ground truth (i.e., $D_{gt} > D_{pred}$) as shown in Table~\ref{tab:smoothness_bias_detailed}. This reveals that current models suffer from an smoothness bias. They fail in volatile domains because they attempt to repair legitimate discontinuities by selecting patches that connect more seamlessly. The error signifies models' failures to grasp the specific physical dynamics of targeting domains and highlights the critical utility of the PZ type question: it acts as a discriminator for temporal fidelity, penalizing models that rely on generic smoothing priors and rewarding those that can capture the specific irregular structural dynamics of the target domain.
\\

\noindent \textbf{Comparison}. 
We analyze model performance on the Comparison task, specifically investigating whether providing explicit domain-level context affects model accuracy. The task requires comparing two input time series, which we test under two conditions: (1) when both series originate from the same domain and (2) when they are from different domains. In both scenarios, the corresponding domain names are provided to the model as textual description.  As shown in Table~\ref{tab:analysis_comparison}, we observe no significant performance difference between the same-domain and different-domain settings across either MC or TF questions. This finding suggests that our Comparison task is domain-invariant. Additionally, combining the results from our analysis on number of topics \& subtopics vs. scores from Table~\ref{tab:subtopic-topic-analysis}, these results indicate that the model's performance is notably stable, which again proves the quality of the proposed dataset. Consequently, to answer correctly, models must reason based on the intrinsic patterns of the time series data itself, rather than relying on the textual context as a simple heuristic. \\

\begin{table}[h]
    \centering
    \scriptsize
    \begin{tabular}{c|l | c c | c c}
        \toprule
        \multirow{2}{*}{\textbf{Group}} & \multirow{2}{*}{\textbf{Model}} & \multicolumn{2}{c|}{\textbf{Same Domain}} & \multicolumn{2}{c}{\textbf{Different Domain}} \\
    
        \cline{3-6} 
        & & MC & TF & MC & TF \\
        \midrule
        
        \multirow{4}{*}{\makecell{Zero\\Shot}}
        & GPT-4.1            & 76.27 & 83.62 & 78.06 & 83.48 \\
        & Gemini-2.5-Flash   & 70.97 & 77.90 & 74.06 & 77.63 \\
        & Qwen3-8B           & 62.99 & 70.67 & 63.54 & 71.60 \\
        & LLaMA3.1-8B        & 49.43 & 67.13 & 50.82 & 68.93 \\
        \midrule 
        
        \multirow{2}{*}{\makecell{Instruction\\Tuning}}
        & Qwen3-8B           & 77.04 & 85.64 & 82.14 & 87.95 \\
        & LLaMA3.1-8B        & 78.02 & 86.32 & 81.24 & 87.35 \\
        \bottomrule 
    \end{tabular}
    \caption{Analysis of Comparison tasks.}
    \label{tab:analysis_comparison}
\end{table}
\subsection{Case Study}
\label{appendix:case_study}
In this category, we analyze selected findings from model outputs, focusing on interesting behaviors observed in our newly proposed Puzzling question type. The insights from this analysis may provide useful implications for future work. Specifically, we examine Puzzling (PZ) questions that models answered incorrectly and present several representative case examples.\\

\begin{figure}[t]
    \centering
    \includegraphics[width=1\linewidth]{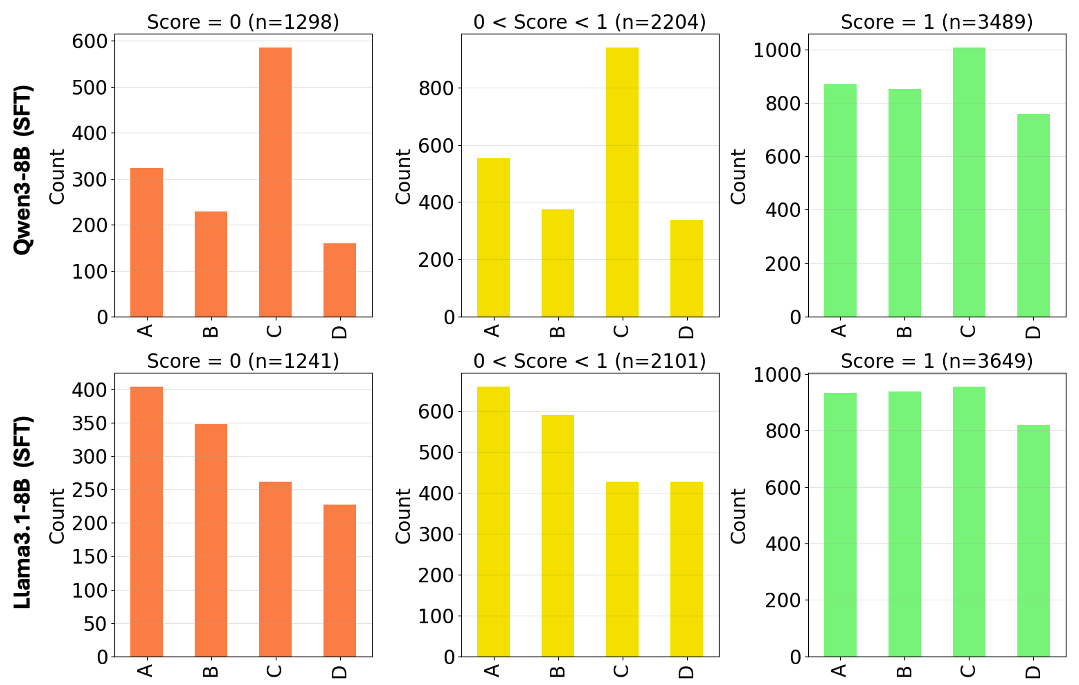}
    \caption{First-letter distribution of model outputs for LLaMA3.1-8B and Qwen3-8B on PZ questions}
    \label{fig:first_letter}
\end{figure}

\noindent \textbf{First Letter Distribution.} 
To explore potential biases in model behavior, we analyzed the distribution of the first letters in model responses for Puzzling questions. Each question provides the first time-series slice and requires reordering of the remaining four. We visualize these distributions in Figure~\ref{fig:first_letter}, using the instruction-tuned LLaMA3.1-8B and Qwen3-8B, whose outputs adhere more strictly to the expected format. Based on the figure, we observe that for both models, the output distributions of questions that received full credit appear approximately uniform, with similar counts across all possible choices. However, for questions that were answered incorrectly or received partial credit, an interesting pattern emerges: Qwen3-8B tends to output the choice C more frequently, whereas LLaMA3.1-8B shows a stronger tendency to output A. \\

\noindent \textbf{Incorrect Output Format.} While models generally demonstrate a strong understanding of the expected response format for common question types such as MC and TF, we further analyzed several examples from the closed-source models GPT-4.1, GPT-4o, and Gemini 2.5 Flash on PZ questions. We include sample responses from each model that highlight unique or unexpected behaviors—cases where the models’ answers do not strictly adhere to the specified instructions for the PZ question format. 

Sample responses from GPT-4o:
\begin{lstlisting}[style=promptstyle]
Question: 'The given time series is the first patch of the sequence.\nBelow are the remaining patches, labeled as A, B, C, and D.\nArrange A, B, C, D in the correct order to reconstruct the original sequence.\n\nPatches:\nA: [-1.6371, 0.2207, 0.9432, 0.6299, 0.7266, 0.1741, -1.2879, -1.9146, 0.2064, 0.7534, 0.5448, 0.4051, 0.4329, -1.5225, -1.86, 0.0201, 0.4759]\nB: [-1.3166, -1.8215, 0.7561, 0.9665, 0.7722, 0.8286, 1.125, -1.262, -1.7866, -1.4097, 0.6872, 1.3175, 0.9611, 1.2673, -0.9737, -1.3918, 0.6585]\nC: [0.2207, 0.4803, 0.5153, -1.3291, -1.7803, 0.398, 0.6684, 0.4141, 0.5305, 0.7472, -1.3524, -1.6084, 0.5636, 0.9396, 0.1759, 0.6039, 0.7579]\nD: [0.4938, 0.6209, -1.3506, -1.7982, 0.2323, 0.7454, -0.1356, 0.6317, 0.6102, -1.3828, -2.1053, 0.3595, 0.86, 0.5752, 0.3998, 0.8241, -1.1912]\n'.

Model Response: 'Input TS -> C -> D -> A -> B'.

Correct Answer: 'D,A,C,B'.
\end{lstlisting}
Sample responses from GPT-4.1:
\begin{lstlisting}[style=promptstyle]
Question: 'The given time series is the first patch of the sequence.\nBelow are the remaining patches, labeled as A, B, C, and D.\nArrange A, B, C, D in the correct order to reconstruct the original sequence.\n\nPatches:\nA: [0.783, 0.918, 0.9904, 1.0754, 1.1464, 1.2123, 1.216, 1.3001, 1.2585, 1.2408, 1.2431]\nB: [-0.0222, -0.1759, -0.3978, -0.5977, -0.6948, -0.8242, -0.9167, -0.9671, -1.0311, -1.1072, -1.1483]\nC: [-1.2175, -1.3207, -1.3356, -1.38, -1.4384, -1.4477, -1.4655, -1.5178, -1.5384, -1.5865, -1.628]\nD: [1.251, 1.2076, 1.1889, 1.1529, 1.0628, 0.9376, 0.8265, 0.6709, 0.4682, 0.3388, 0.1427]\n'.

Model Response: 'B, C, [-0.1936, -0.1581, -0.0984, -0.0119, 0.0745, 0.1174, 0.2594, 0.3837, 0.4481, 0.6023, 0.7041], A, D'.

Correct Answer: 'A,D,B,C'. 
\end{lstlisting}
Sample responses from Gemini-2.5-Flash:
\begin{lstlisting}[style=promptstyle]
Question: 'The given time series is the first patch of the sequence.\nBelow are the remaining patches, labeled as A, B, C, and D.\nArrange A, B, C, D in the correct order to reconstruct the original sequence.\n\nPatches:\nA: [0.9363, -0.4796, 0.7003, -0.0077, -1.4236, -1.4236, -1.1876, -1.1876, 0.4643, -0.2436, -0.4796, 0.9363, -0.7156, -0.2436, -2.1316, 0.2283, 0.4643, -0.7156, 0.9363, -0.2436, -0.2436, 0.2283, 2.3523, -0.2436, 0.2283, -0.9516, 0.7003, -1.6596, 0.2283, 0.4643, -0.4796, 0.7003, 0.7003, -0.2436, -0.0077, -0.9516, -1.1876]\nB: [0.2283, 0.9363, -0.2436, 1.6443, 0.2283, 0.7003, 0.4643, -0.9516, 0.2283, -0.0077, -0.2436, -0.7156, 0.7003, -0.7156, 0.4643, -0.7156, -0.0077, 3.0603, 0.7003, -1.4236, 0.9363, -0.7156, -1.6596, -0.4796, -1.1876, 0.7003, -0.2436, -0.9516, 0.7003, 0.4643, -0.4796, -1.8956, -0.7156, 1.6443, 1.8803, -1.1876, -0.0077]\nC: [-0.2436, -1.4236, -0.7156, -1.1876, -0.0077, -0.0077, -0.0077, 1.4083, -0.2436, -0.7156, -0.9516, 0.2283, 0.4643, 1.6443, 0.9363, 3.7682, 0.4643, -0.7156, 0.2283, -0.0077, -0.0077, -0.7156, -1.6596, 1.6443, -0.4796, -0.4796, -1.6596, -2.1316, -0.7156, 0.4643, 2.5883, 2.5883, 0.2283, 1.1723, -1.6596, 1.4083, 1.4083]\nD: [-0.0077, -0.0077, -1.4236, -0.4796, -0.2436, 0.4643, -0.2436, -1.6596, 1.4083, -1.4236, -1.1876, -0.7156, 0.7003, 1.4083, -0.0077, -0.2436, 0.9363, -0.2436, -0.2436, 1.1723, 1.1723, 0.4643, -0.4796, -0.2436, -0.0077, -0.0077, 0.2283, -1.1876, -0.4796, -0.4796, 0.2283, 0.4643, 0.2283, -0.7156, 0.9363, -0.0077, -0.9516]\n'.

Model Response: '{\n "answer": "The provided time series patches do not exhibit any standard prefix-suffix overlaps. A common approach for reconstructing a sequence from patches is to identify the longest overlapping segments between the end of one patch and the beginning of another. After performing a comprehensive search for such overlaps (ranging from 1 to 37 elements)........the default alphabetical order is chosen."\n}'.

Correct Answer: 'A,B,C,D'.
\end{lstlisting}

Based on the sample responses, we observe that the Puzzling question type often causes models to struggle in following the question instructions. This finding highlights the continued importance of developing and designing more reasoning-based time-series questions beyond traditional formats. Such question types better evaluate a model’s ability to perform complex reasoning and manipulate temporal information, providing a more comprehensive assessment of time-series understanding.

\newpage
\section{Human Evaluation}
\label{appendix:human_evaluation}
\begin{figure*}[h]
    \centering
    \includegraphics[width=0.9\linewidth]{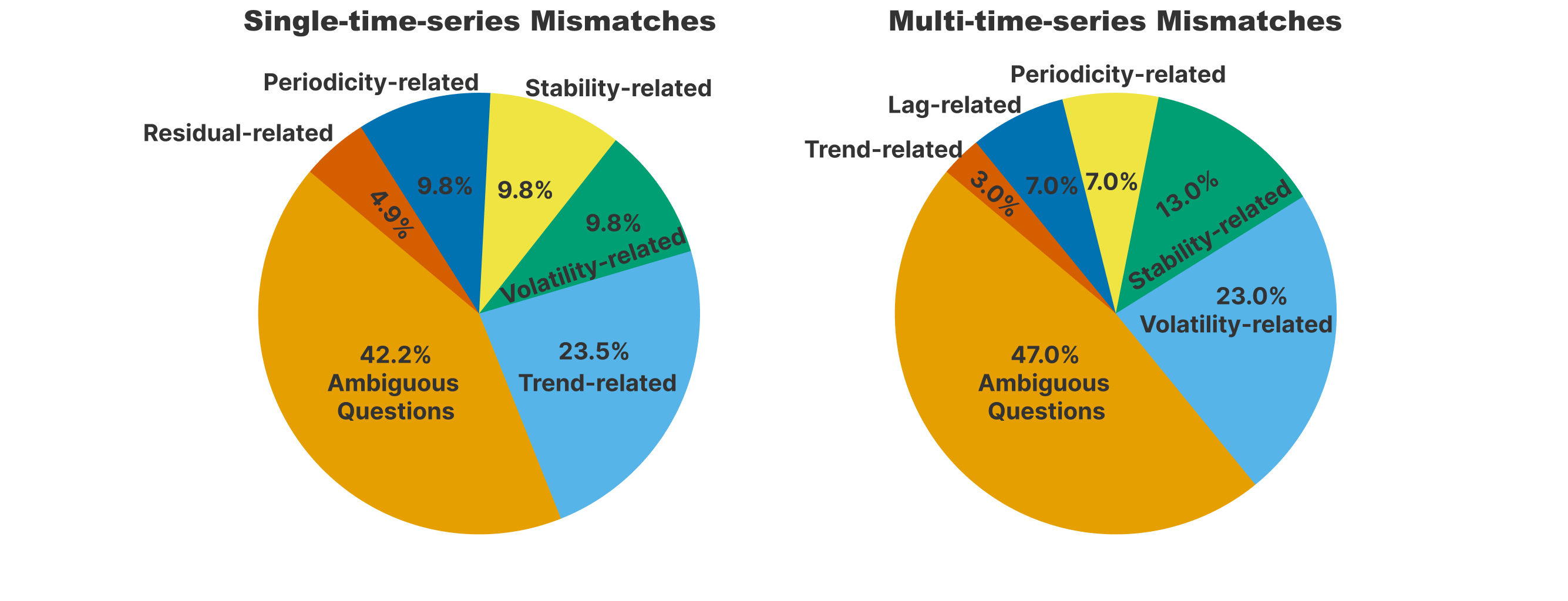}
    \caption{Human explanations for answer mismatches in \trqa}

    \label{fig: mismatch}
\vspace{-5mm}
\end{figure*}

We further examine annotators’ explanations in cases of disagreement. In the single-series benchmark, the largest source of mismatches is ambiguous questions (43\%). Among well-formed cases, 24\% involve trends, while volatility-, stability-, and periodicity-related issues each account for 10\%. A small fraction (5\%) reflects residual annotator uncertainty. In the multi-series benchmark, mismatches are more strongly tied to stochastic properties: volatility-related issues dominate (23\%), followed by stability (13\%). Periodicity- and lag-related issues each contribute 7\%, while trend-related mismatches are rare (3\%). Nearly half of the disagreements (47\%) again arise from ambiguous questions, underscoring the greater interpretive difficulty of the multi-series setting. (See Figure~\ref{fig: mismatch})

\end{document}